\documentclass[10pt,journal,compsoc,]{IEEEtran}
\usepackage{hyperref}
\hypersetup{colorlinks, citecolor=blue}
\usepackage{amsmath,amssymb,amsfonts}
\usepackage{algorithm}
\usepackage{algpseudocode}
\usepackage{subfigure}
\usepackage{graphicx}
\usepackage{textcomp}
\usepackage{wrapfig}
\usepackage{csquotes}
\usepackage[modulo]{lineno}
\usepackage{booktabs}
\usepackage{xcolor}
\usepackage{ragged2e}
\usepackage{caption}

\ifCLASSOPTIONcompsoc

  \usepackage[nocompress]{cite}
\else
  \usepackage{cite}
\fi
\ifCLASSINFOpdf
  
\else
  
\fi
\usepackage[]{footmisc}
\begin{document}
\title{Meta Attentive Graph Convolutional Recurrent Network for Traffic Forecasting}
\author{Adnan Zeb, Yongchao Ye,~\IEEEmembership{Student Member,~IEEE}, ~Shiyao Zhang,~\IEEEmembership{Member,~IEEE},
        and~James J. Q. Yu,~\IEEEmembership{Senior Member,~IEEE}
\thanks{}}
\markboth{Journal of \LaTeX\ Class Files,~Vol.~14, No.~8, August~2015}%
{Shell \MakeLowercase{\textit{et al.}}: Bare Demo of IEEEtran.cls for Computer Society Journals}

\IEEEtitleabstractindextext{%
\justify 
\begin{abstract}
Traffic forecasting is a fundamental problem in intelligent transportation systems. Existing traffic predictors are limited by their expressive power to model the complex spatial-temporal dependencies in traffic data, mainly due to the following limitations. Firstly, most approaches are primarily designed to model the local shared patterns, which makes them insufficient to capture the specific patterns associated with each node globally. Hence, they fail to learn each node's unique properties and diversified patterns. Secondly, most existing approaches struggle to accurately model both short- and long-term dependencies simultaneously. In this paper, we propose a novel traffic predictor, named Meta Attentive Graph Convolutional Recurrent Network (MAGCRN). MAGCRN utilizes a Graph Convolutional Recurrent Network (GCRN) as a core module to model local dependencies and improves its operation with two novel modules: 1) a  Node-Specific Meta Pattern Learning (NMPL) module to capture node-specific patterns globally and 2) a Node Attention Weight Generation Module (NAWG) module to capture short- and long-term dependencies by connecting the node-specific features with the ones learned initially at each time step during GCRN operation. Experiments on six real-world traffic datasets demonstrate that NMPL and NAWG together enable MAGCRN to outperform state-of-the-art baselines on both short- and long-term predictions. 
\end{abstract}
}

\maketitle
\IEEEdisplaynontitleabstractindextext
\ifCLASSOPTIONpeerreview
\begin{center} \bfseries \end{center}
\fi
\IEEEpeerreviewmaketitle

\IEEEraisesectionheading{\section{Introduction}\label{sec:introduction}}

\IEEEPARstart{W}{ith} the continued growth in population and travels demand, the transportation infrastructure in cities confronts severe challenges in traffic management. This problem has been widely acknowledged across the globe and has called for an effective intelligent transportation system (ITS) to establish a safe transportation network with efficient utilization of available resources \cite{LinASO, ZhangDataDrivenIT, Yu2022GraphCF, Yu2021LongTermUT}. ITS has several essential components, among which this work focuses on one of the fundamentals: traffic forecasting. Traffic forecasting aims at predicting future traffic conditions based on historical traffic state observations. These predictions are analyzed for improving traffic management and congestion prevention strategies \cite{Jiang2021DLTraffSA, Jia2018DataDC, Kong2013UTNModelBasedTF, Zhu2019BigDA, Snyder2019STREETSAN}. Due to its great practical value, traffic forecasting has raised the interest of both academic and industrial communities in making efforts to develop reliable prediction frameworks.  

Traffic data comprises readings recorded continuously at different time steps via sensors deployed on different road segments. Hence, traffic data exhibits non-linear spatial and temporal dependencies among the sensors, which must be modeled for accurate traffic forecasting. Many techniques have been developed for modeling traffic dependencies in the past decades, ranging from the classical time-series models, such as Integrated Moving Average (ARIMA) \cite{Williams2003ModelingAF}, Vector Auto-Regressive (VAR) \cite{Zivot2003VectorAM}, and Support Vector Regression (SVR) \cite{Drucker1996SupportVR}, to the recent deep neural networks (DNNs). In DNNs, a class of techniques, namely spatial-temporal graph neural networks (STGNNs), have offered substantial abilities in capturing complex inter-dependencies in traffic data. STGNNs represent a traffic network as a graph where each node is a monitoring sensor. Then, they mainly utilize graph convolution \cite{Kipf2017SemiSupervisedCW, Velickovic2018GraphAN} to capture spatial dependency and recurrence \cite{Hochreiter1997LongSM}, convolution \cite{LeCun1998GradientbasedLA}, or attention \cite{Vaswani2017AttentionIA} to model temporal dependencies among the sensors. Based on the temporal dependency modeling mechanism, STGNNs are primarily categorized into recurrence-based \cite{Li2018DiffusionCR, Bai2020AdaptiveGC, Zhao2020TGCNAT, Wang2020TrafficFP,chen2021z}, convolution-based \cite{Wu2019GraphWF, Huang2020LSGCNLS, Yu2018SpatioTemporalGC, Guo2019AttentionBS, Song2020SpatialTemporalSG, li2021spatial}, and attention-based methods \cite{Guo2019AttentionBS, huang2019dsanet, lan2022dstagnn}. Although these STGNNs have previously demonstrated promising achievements in traffic forecasting, they vastly suffer from the following critical limitations.    
\begin{figure}[h]
	\centering
		\includegraphics[width=\linewidth]{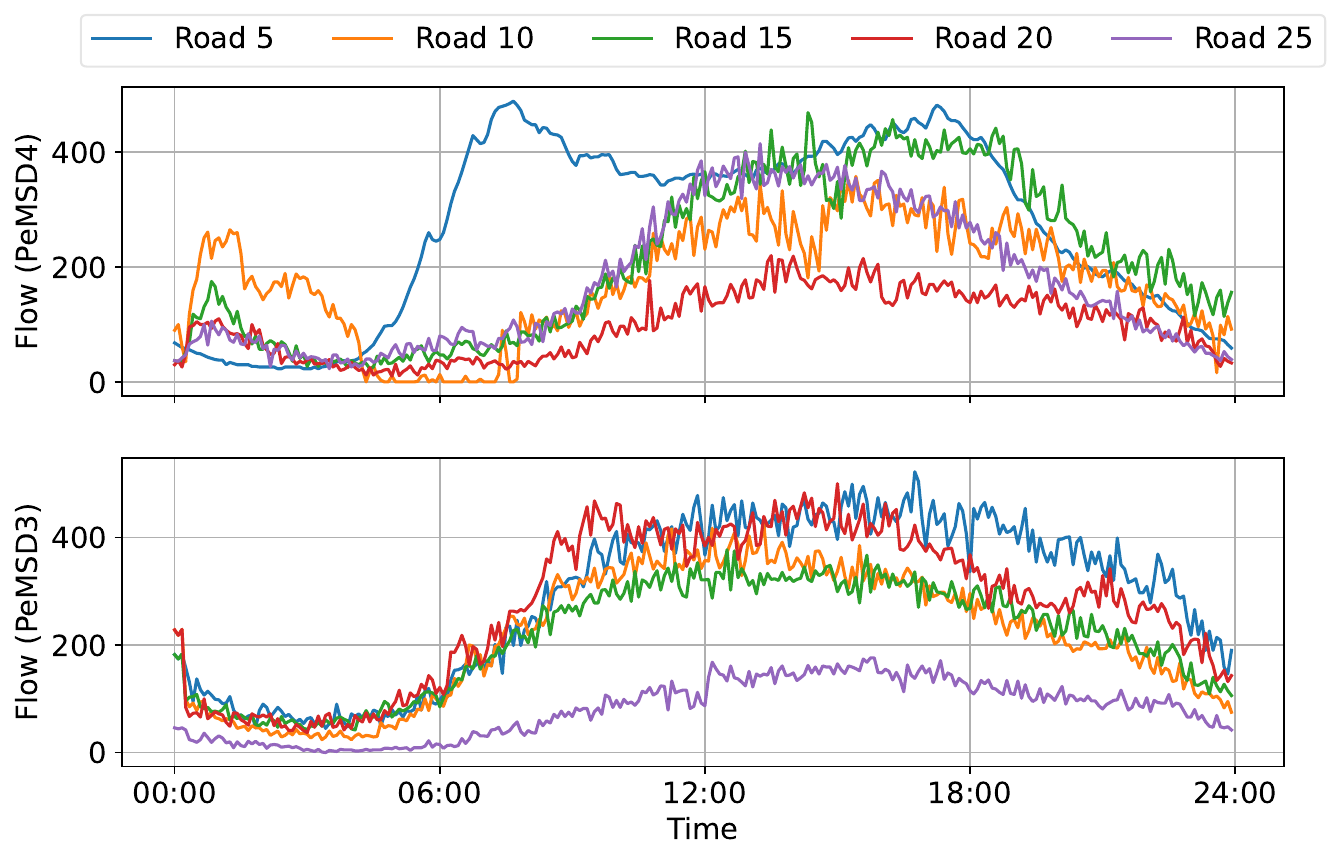}
        \caption{Flow values for five roads in the PeMSD3 and PeMSD4 datasets. Each road fluctuates in a different range and shows distinct patterns from the other nodes. In addition, each road reveals distinct patterns at distinct time steps but follows a specific increasing/decreasing order in the morning, noon, and evening.}
        \label{fig: fig1}
\end{figure}

 First, STGNNs still lack the ability to make accurate short- and long-term forecasts simultaneously. Recurrence-based methods employ graph convolutional recurrent networks (GCRNs), which seem a natural fit for modeling continuous traffic data. However, these GCRNs are limited by their internal cyclic operation as it considers the temporal information to be sequentially dependent, and they also struggle with maintaining the information for long sequences, which restricts GCRNs' ability to local temporal patterns and short-range dependencies \cite{Li2022DynamicGC, Tang2020JointMO, Yao2018DeepMS, Flunkert2020DeepARPF}. Convolution-based methods are widely known as inflexible because of their implicit temporal modeling mechanism, which causes the time steps to be invisible \cite{Li2022DynamicGC}. In addition, the limited receptive fields of the convolutional kernels make it harder to capture the long-range dependencies \cite{Guo2019AttentionBS}. Some efforts have been made via implementing temporal versions of the standard convolution \cite{vanwavenet}, \cite{bai2018empirical}. These methods leverage dilated convolution \cite{Yu2015MultiScaleCA} with large receptive fields but require multiple convolutional layers to connect a pair of positions in a given sequence, which unfavorably leads to over-fitting on global information and losing the local ones \cite{Vaswani2017AttentionIA}.

Attention-based methods are inspired by convolution-based methods, except for replacing convolution with attention to aggregate temporal information. These methods have an advantage over convolution- and recurrence-based methods in modeling the dependencies regardless of their distance in the given input sequence, which enables them to model the long-range dependencies \cite{Guo2019AttentionBS, lan2022dstagnn}. Since these methods do not contain recurrence, they utilize positional encoding as a substitute to insert positional information and make use of the order of the sequence. However, it has been recently shown for multi-step retail and volatility forecasting that positional encoding is insufficient to capture sequence order in time-series data, where multiple complex patterns exist across the time steps \cite{Lim2019TemporalFT}. Traffic data has identical features to retail and volatility data. Specifically, the traffic conditions recorded via a sensor at different time steps are mainly different due to time-specific conditions. As shown by an example in Fig. \ref{fig: fig1}, each road exhibits distinct flow patterns at distinct time steps but follows a specific increasing/decreasing order in the sequence. For example, the flow values increase after 6:00 while decreasing after 18:00. In such circumstances, positional encoding may lack the ability to accurately utilize the order of the sequence and make a precise future prediction.  
Second, existing STGNNs are primarily designed to model the shared patterns among all nodes. This shared design makes existing methods insufficient in capturing node-specific patterns. In fact, each node carries diversified patterns owning to distinct attributes, such as location, time, and weather. A demonstration of distinct patterns exhibited by nodes can also be visually observed in Fig. \ref{fig: fig1} along with a description in the caption. Although some existing works have explicitly mentioned this limitation of shared design, none has attempted specifically to learn the node-specific patterns except a recent GCRN \cite{Bai2020AdaptiveGC}, which learns node adaptive parameters sequentially in a cyclic manner across each time step during recurrence operation. As mentioned earlier, recurrence extracts 1) the patterns locally and 2) struggles to retain the patterns when the sequence is longer, potentially degrading the effectiveness of learning node adaptive parameters and limiting the model to short-range sequences.

To address the abovementioned problems, we propose a novel traffic forecasting model, named Meta Attentive Graph Convolutional Recurrent Network (MAGCRN). MAGCRN is built upon the GCRN framework. Specifically, MAGCRN leverages GCRN as a core module and improves it with two additional novel modules for traffic forecasting: 1) a Node-Specific Meta Pattern Learning (NMPL) module that leverages the node parameters of the GCRN module to generate convolutional filters and then uses them on the output of the last GCRN layer to extract the node-specific features globally for each graph node 2) a Node Attention Weight Generation (NAWG) module, which utilizes the entire sequence output of the GCRN module to generate attention weights for the node-specific features extracted in the previous module. The former module handles the first recurrence limitation by updating the local node parameters of the GCRN module with node-specific global patterns to learn a mixed representation of both. The latter module handles the second recurrence limitation by enriching the node-specific features with patterns learned initially at each GCRN layer for each time step. Both of these settings improve the temporal modeling capability of the GCRN module. 

Our main efforts are summarized as follows:
\begin{itemize}
    \item We design a novel module NMPL that extracts node-specific global patterns for enriching local node parameters to learn a mixed representation of both local and global temporal patterns.

    \item We develop a cross-attentive module NAWG that generates attention weights from the features of all time steps and assigns those weights to the node-specific features to enrich them with short- and long-range patterns. 

    \item We combine NMPL and NAWG with GCRN and propose a unified traffic predictor MAGCRN that can simultaneously model local, global, short-, and long-range dependencies in a traffic network.

    \item  We conduct comprehensive experiments on four traffic flow and two traffic speed datasets. Experimental results demonstrate that the proposed MAGCRN offers more accurate short- and long-term prediction on both small and large datasets than a large pool of recent state-of-the-art traffic forecasting models.    
\end{itemize}

The remainder of this paper is structured in the following sections: Sections 2 and 3 discuss the related work and traffic forecasting problem, respectively. Section 4 presents the proposed model. Section 5 presents the experimental settings and results on traffic prediction. Section 6 concludes the writing with final remarks and future directions. 
\section{Related Work}
\label{sec:secrelatedwork}
Traffic forecasting is one of the principal research directions in ITS that has recently attracted significant attention due to its great application value. Earlier works in this domain include time-series analysis methods such as ARIMA \cite{Williams2003ModelingAF}, VAR \cite{Zivot2003VectorAM}, and SVR \cite{Drucker1996SupportVR}. These methods are either based on linear assumption or require hand-crafted features, where both cases are limited in capturing the fine-grained spatial-temporal dependencies in complex traffic networks. This limitation of classical machine learning methods brought the attention of researchers towards developing a bunch of deep techniques called STGNNs. STGNNs mainly employ GCNs to model spatial dependencies but vary in terms of modeling temporal dependencies. Based on the temporal modeling mechanism, STGNNS are broadly categorized into recurrence-based, convolution-based, and attention-based methods.

\subsection{Recurrence-based Methods}
Recurrence-based methods employ recurrent neural networks (RNNs), long-short-term memory networks (LSTMs), or gated recurrent units (GRUs) to store historical information on the given sequence. In this category, we discuss the recent state-of-the-arts, which are mostly GRUs-based that replace the linear projections in recurrent units with graph convolution and are called graph convolutional recurrent networks (GCRNs). TGCN \cite{Zhao2020TGCNAT} is a representative model of this type, which replaces linear projections in recurrent units with standard graph convolution \cite{Kipf2017SemiSupervisedCW} to model spatial-temporal correlations. DCRNN \cite{Li2018DiffusionCR} is another GCRN that embeds diffusion graph convolution in recurrent units to model spatial-temporal dependencies. STGNN \cite{Wang2020TrafficFP} implements an additional transformer layer \cite{Vaswani2017AttentionIA} on the output of the recurrent units to embed long-term temporal dependencies. AGCRN \cite{Bai2020AdaptiveGC} additionally exploits the learnable node embeddings with a weight pool matrix in the standard graph convolution to learn node adaptive parameters during the recurrence operation. Z-GCNETs \cite{chen2021z} integrates zigzag persistence with time-aware graph convolution in the recurrent units to enhance the predictive performance. Although these GCRNs fit naturally on sequential traffic data, the recurrent units gradually forget historical information as the number of recurrence steps increases, which subsequently causes information loss and limits the predictive performance \cite{Bengio1994LearningLD, Pascanu2012OnTD}.

\subsection{Convolution-based Methods}
Convolution-based methods operate in a similar way to the recurrence-based methods, except replacing RNNs with temporal convolutional networks (TCNs). In particular, they integrate temporal convolution with graph convolution to model spatial-temporal dependencies. STGCN \cite{Yu2018SpatioTemporalGC} is one of the earlier works in this category that combines gated temporal convolution with graph convolution to capture spatial-temporal dependencies. STSGCN \cite{Song2020SpatialTemporalSG} extends STGCN with a multiple-module mechanism to detect multiple dynamics in the given input data. Another prominent example of these methods is the well-known Graph WaveNet \cite{Wu2019GraphWF}, which leverages 1D dilated convolution to make exponentially large receptive fields for modeling long-range sequences. LSGCN \cite{Huang2020LSGCNLS} introduces a new graph attention network, incorporating it with graph convolution and gated linear convolution to model spatial and temporal correlations. STFGNN \cite{li2021spatial} introduces a new spatial-temporal fusion module and operates it in parallel with a gated dilated convolution module to capture local and global dependencies. 

These convolution-based methods have empirically shown to be a strong contender to the recurrence-based methods. However, in practice, these methods usually tend to compose large receptive fields for long-range dependencies that require implementing multiple convolutional layers to connect a pair of positions in the given sequence. This process makes these models overfit on the global information.  

\subsection{Attention-based Methods}
Attention-based methods follow the same architectural design of recurrence- and convolution-based methods. However, these methods consider that the spatial and temporal dependencies dynamically change over time. ASTGCN \cite{Guo2019AttentionBS} is a prominent model of this class that implements spatial and temporal attention layers. The spatial attention layer dynamically updates the graph adjacency and conducts graph convolution, while the temporal attention layer computes the weighted sum of nodes' historical states and assigns them attention scores. MSTGCN \cite{Guo2019AttentionBS} is a variant of ASTGCN that drops the spatial-temporal attention module to validate its importance in capturing spatial-temporal dependencies in traffic data. DSANet \cite{huang2019dsanet} implements two convolutional models in parallel to capture a mixture of local and global temporal patterns. The outputs are then provided to a self-attention component to model dependencies between multiple series. DSTAGNN \cite{lan2022dstagnn} is a recent state-of-the-art attention-based model that designs a novel dynamic spatial-temporal patterns aware graph to replace the traditional static and adaptive ones used by the previous methods in the graph convolution operation. A primary advantage of the attention-based methods is their receptive field that globally aggregates information and allows the modeling of long-range dependencies. However, due to the inability of positional encoding to accurately utilize the order of the input sequence \cite{Lim2019TemporalFT}, these methods may potentially overlook certain features, such as the increasing/decreasing patterns in continuous time-series traffic data.   

In contrast to these methods, the proposed MAGCRN fits all the abovementioned categories. In particular, our model uses recurrent units that learn local temporal patterns and utilize the sequence order to capture the increasing/decreasing patterns. In addition, MAGCRN leverages a meta-learning convolutional technique hypernetwork, which extracts the node-specific patterns globally and updates the local node parameters with the global patterns to learn mixed patterns of both local and global dependencies. Furthermore, it utilizes a cross-attention module to generate attention weights from the features learned at each GCRN layer and assign those weights to the node-specific features to support short- and long-range dependencies.  
\begin{figure*}[t]
	\centering
		\includegraphics[width=.99\linewidth]{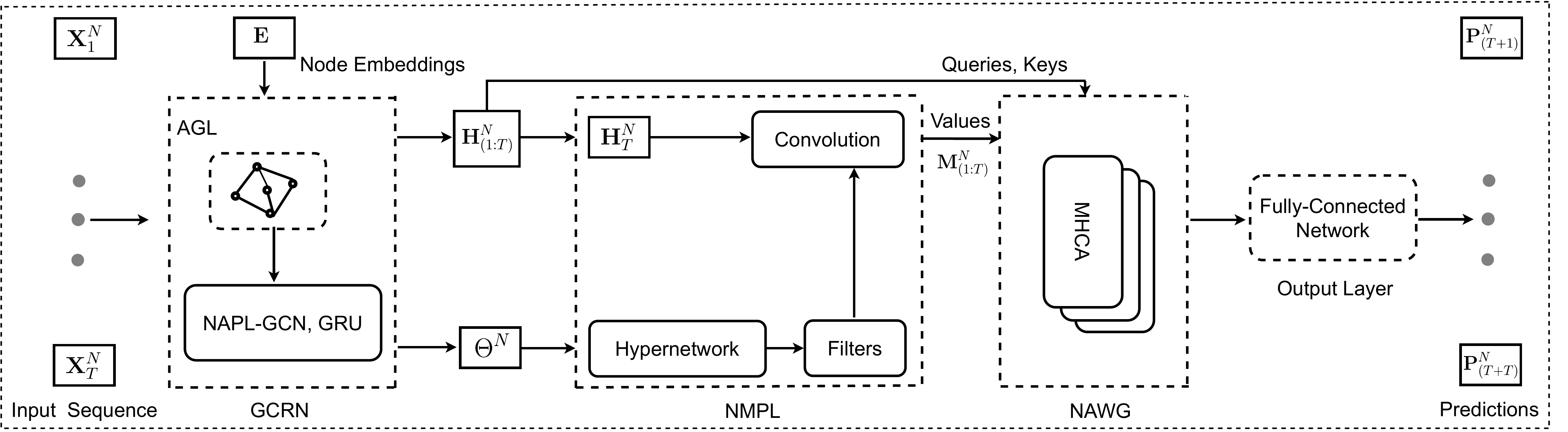}
	\caption{Architecture of the proposed MAGCRN.}  
        \label{fig: fig4}
\end{figure*}
\section{Problem Definition}
A traffic network is typically viewed as a graph where nodes correspond to traffic sensors/stations linked via hidden spatial-temporal relationships. A traffic graph can be directed, undirected, weighted, or unweighted based on the raw data and the graph generation schemes. Let $\mathcal{G} = \left( \mathcal{V}, \mathcal{E} \right)$ denotes a traffic network. Then,  $\mathcal{V}$ is the set of nodes, and $\mathcal{E}$ is the set of edges, symbolizing the connectivity structure between nodes in the given graph $\mathcal{G}$. Let $\textbf{A} \in \mathbb{R}^{N \times N }$ be a geometric or binary adjacency matrix built upon the graph edges $\mathcal{E}$ with topological information of the traffic network. Besides, $\textbf{X}^{N}_{t} = \{x^{1}_{t}, x^{2}_{t},..., x^{N}_{t}\} \in \mathbb{R}^{N \times 1}$ represents traffic values of $N$ nodes at time step $t$. Then, the traffic values for $T$ time steps can be written as a matrix $\textbf{X}^{N}_{(1:T)} = \{\textbf{X}^{N}_{1}, \textbf{X}^{N}_{2}, ..., \textbf{X}^{N}_{ T}\}$. For a given historical traffic state sequence of $T$ time steps, the objective is to forecast the future traffic state values for the next $T$ steps:

\begin{subequations}
\begin{align}
\textbf{P}^{N}_{(T+1:T+T)} &= \mathcal{F}_{\omega}(\textbf{X}^{N}_{(1:T)}, \mathcal{G}), \\
 \{\textbf{P}^{N}_{T + 1}, \textbf{P}^{N}_{T + 2}, \hdots, \textbf{P}^{N}_{T + T}\} 
&= \mathcal{F}_{\omega}(\textbf{X}^{N}_{1}, \textbf{X}^{N}_{2},  \hdots, \textbf{X}^{N}_{T},\mathcal{G}),
\end{align}     
\end{subequations}
where $\mathcal{F}$, $\textbf{P}^{N}_{(T+1:T+T)}$, and $\omega$ symbolize the forecasting function, the matrix of predicted values, and all the learnable model parameters, respectively.
\section{Meta Attentive Graph Convolutional Recurrent Network}
In this section, we introduce the proposed MAGCRN for traffic forecasting. First, we provide a generic overview of the model architecture and discuss the structural and operational improvements it brings over existing STGNNs. Second, we discuss the detailed implementations and operations of the core modules. 

\subsection{Model Architecture}
There are four core modules in MAGCRN: an adaptive graph learning (AGL) module, a graph convolutional recurrent network (GCRN) module, a node-specific meta pattern learning (NMPL) module, and a node attention weight generation (NAWG) module. While AGL is designed for learning graph structure adaptively from the node embeddings $\textbf{E}$, GCRN is developed explicitly for modeling spatial-temporal dependencies in the traffic sequence $\textbf{X}^{N}_{(1:T)}$, which are the two fundamental objectives of many STGNNs for traffic forecasting. AGL works as an integral operation of the GCRN module. NMPL and NAWG are two novel additions of MAGCRN to operate over GCRN.

In NMPL, we employ a meta-learning technique hypernetwork \cite{Hahypernetwork, Balazevic2018HypernetworkKG} to develop convolutional filters from the node parameters $\Theta^{N}$ learned during GCRN operation. Specifically for each node, we develop $T$ filters corresponding to $T$ horizons since we aim to generate predictions as a whole rather than auto-regressively through $T$ steps. These convolutional filters are then convolved with the output of the last GCRN layer $\textbf{H}^{N}_{T}$, where each filter is responsible for learning node-specific patterns. The primary advantage of this component is enabling the extraction of node-specific patterns globally from the input sequence, which is useful in learning the diversified trends exhibited by each node.

In the NAWG module, we employ the multihead cross-attention (MHCA) mechanism, which receives input from both the GCRN and NMPL modules. NAWG treats the full sequence output of the GCRN $\textbf{H}^{N}_{(1:T)}$ as the queries and keys and the NMPL output $\textbf{M}^{N}_{(1:T)}$ as the values for MHCA operation. It then generates attention weights from the queries and keys and assigns them to the values, i.e., the node-specific features extracted in the NMPL module. Fig. \ref{fig: fig4} demonstrates the detailed architecture of our model that how NMPL and NAWG modules collaborate and operate over the GCRN module. Finally, the output layer employs a fully-connected layer to map the final output to the corresponding scalar predictions. It is worth noting that MAGCRN utilizes the same input information as the existing STGNNs, without including any additional information. In more detail, the core operation of each module is discussed in the following.

\subsection{Adaptive Graph Learning (AGL)}
Graph structure is vital for traffic forecasting since it contains topological information required for modeling spatial dependencies in a traffic network. STGNNs mainly implement two schemes to utilize the graph structure: 1) pre-defined \cite{Li2018DiffusionCR, Zhao2020TGCNAT}, and 2) adaptive \cite{Bai2020AdaptiveGC, Wang2020TrafficFP, Wu2019GraphWF}. The pre-defined structures are built in advance upon the network edges, requiring distance metrics to compute the geographical distance between the nodes or the similarity between their attributes. However, it has been demonstrated in recent works that the pre-defined structures lack the ability to capture the dynamic nature of the traffic networks. Hence, the adaptive scheme emerged as a productive alternative to the previous one. Adaptive graphs dynamically learn the hidden spatial dependencies in an end-to-end fashion during model training and hence do not rely on any pre-defined structure. Inspired by the adaptive graph structures \cite{Bai2020AdaptiveGC, Wang2020TrafficFP, Wu2019GraphWF}, our model defines a single embedding matrix $\textbf{E} \in \mathbb{R}^{N \times C}$ for graph learning where each row is a randomly initialized $C$-dimensional feature vector of a node. Hence,  the spatial dependency between nodes is measured by computing the similarity between their feature vectors: 
\begin{equation}
    \textbf{A} =  \phi(\sigma(\textbf{E} \cdot \textbf{E}^{\intercal})),
    \label{eq: eq3}
\end{equation}
where $\phi$, $\sigma$, and $\intercal$ denote the softmax function, the activation function ReLU, and the transpose operator, respectively. Here, the softmax is used to normalize the adaptive matrix.  

\subsection{Graph Convolutional Recurrent Network (GCRN)}
\label{sec:sectionGCRN}
In the previous section, we discussed the AGL module, which operates as an integral component of the GCRN module. In addition, standard GCRNs typically comprise a GCN and an RNN for modeling the spatial and temporal dependencies, respectively. GCNs implement graph convolution over nodes adjacency and feature matrix to aggregate information from the neighboring nodes and fuse them with each corresponding central node. This process allows a GCN to capture the spatial dependencies among nodes in a given traffic network. The graph convolution in a standard GCN is mathematically defined as follows:
\begin{equation}
    \textbf{Z}^{N}_{t} = \sigma (\hat{\textbf{A}}\textbf{X}^{N}_{t}\textbf{W} + b) ,
    \label{eq:eq3}
\end{equation}
where $\hat{\textbf{A}} = \textbf{A} + \textbf{I} \in \mathbb{R}^{N \times N}$ is the adjacency matrix with added identity matrix $\textbf{I}$ for self-loops, $\textbf{X}^{N}_{t} \in \mathbb{R}^{N \times D}$ is the feature matrix at time step $t$, and $\textbf{W}$ is a $D \times D^{'}$-dimensional weight matrix. Eq. (\ref{eq:eq3}) represents a single graph convolutional layer. However, several layers are typically implemented and stacked on top of others. GCRNs usually employ different versions of GCNs, among which we use NAPL-GCN reported in \cite{Bai2020AdaptiveGC}, where NAPL stands for node adaptive parameter learning. In this technique, the node embedding matrix $\textbf{E} \in \mathbb{R}^{N \times C}$ is included in the graph convolution. In addition, it replaces the weight matrix $\textbf{W} \in \mathbb{R}^{D \times D^{'}}$ and the scalar bias $b$ with a weight pool matrix $\textbf{W} \in \mathbb{R}^{C \times D \times D^{'}}$ and a bias pool matrix $\textbf{b} \in \mathbb{R}^{C \times D^{'}}$, respectively. Furthermore, it computes a product of the node embedding matrix with the weight pool and bias matrices, i.e., $\Theta^{N} = \textbf{E}. \textbf{W} \in \mathbb{R}^{N \times D \times D^{'}}$, and $\Omega^{N} = \textbf{E} . \textbf{b} \in \mathbb{R}^{N \times D^{'}}$ before the graph convolution. This process allows it to extract node parameters from the shared weight pool according to the relevant node embeddings. NAPL-GCN defines the graph convolution as follows: 
\begin{equation}
    \textbf{Z}^{N}_{t} = \sigma (\hat{\textbf{A}}\textbf{X}^{N}_{t} \Theta^{N} + \Omega^{N}). 
\end{equation}

To capture temporal dependencies, we combine the top-performing architecture GRU with NAPL-GCN in the GCRN module. The GRU operation \cite{Bai2020AdaptiveGC} is mathematically defined as follows: 
\begin{subequations}
\begin{align}
    \textbf{Z}^{N}_{t} &= \sigma(\hat{\textbf{A}}[\textbf{X}^{N}_{t}, \textbf{H}^{N}_{t-1}] \Theta_{Z}^{N} + \Omega_Z^{N}),\\ 
    \textbf{R}^{N}_{t} &= \sigma(\hat{\textbf{A}}[\textbf{X}^{N}_{t}, \textbf{H}^{N}_{t-1}] \Theta_{R}^{N} + \Omega_R^{N}),   \\ 
    \hat{\textbf{H}}^{N}_{t} &= \tanh(\hat{\textbf{A}}[\textbf{X}^{N}_{t}, \textbf{R}^{N}_{t} \odot \textbf{H}^{N}_{t-1}] \Theta_{\hat{H}}^{N} + \Omega_{\hat{H}}^{N}), \\ 
    \textbf{H}^{N}_{t} &= \textbf{Z}^{N}_{t} \odot \textbf{H}^{N}_{t-1} + (1 -\textbf{Z}^{N}_{t}) \odot \hat{\textbf{H}}^{N}_{t},
\end{align}
\end{subequations}
where $\textbf{Z}$ and $\textbf{R}$ are reset and update gates, respectively. $[\textbf{X}^{N}_{t}, \textbf{H}^{N}_{t-1}]$ and $\textbf{H}^{N}_{t}$ denote input and output at time step $t$, respectively. Operators $\odot$ and $[,]$ represent the dot product and concatenation, respectively. Precisely, the overall operation of the GCRN module for all time steps $T$ can be mathematically written as follows:  
\begin{equation}
\begin{split}
    \textbf{H}^{N}_{(1:T)} = &\, \text{GRU}(\text{NAPL-GCN}(\hat{\textbf{A}}, [\textbf{X}^{N}_{(1:T)}, \\
    &\, \textbf{H}^{N}_{(1-1:T-1)}]; \Theta^{N}; \Omega^{N}); \Phi),
    \label{eq:eq6}
\end{split}
\end{equation}
where $\Phi$ denotes GRU parameters.

\subsection{Node-Specific Meta Pattern Learning (NMPL)}
\label{sec:sectionNMTM}
In the previous section, we discuss that the GCRN module relies solely on the node embeddings and weight pool to compute the node adaptive parameters $\Theta^{N} = \textbf{E}. \textbf{W} \in \mathbb{R}^{N \times D \times D^{'} }$ at each time step during the recurrence operation. Apparently, this technique determines the characteristics of the time-invariant features since both $\textbf{E} \in \mathbb{R}^{N \times C}$ and $\textbf{W}\in \mathbb{R}^{C \times D \times D^{'}}$ are shared among all time steps. However, it ignores the fact that real-world traffic networks are more complex, where each node exhibits highly heterogeneous patterns, and recurrence tends to capture only the local dependencies since it considers the temporal patterns to be sequentially dependent across the time steps. This limitation of recurrence motivates us to implement an additional component for learning the node adaptive parameter matrix $\Theta^{N}$ and enriching it with node-specific patterns extracted globally from the input signals.

To handle the abovementioned problem, MAGCRN first implements the NMPL module. In the feed-forward pass, NMPL receives two matrices from the GCRN module: 1) the node adaptive parameter matrix $\Theta^{N} \in \mathbb{R}^{N \times DD^{'}}$ (reshaped from $N \times D \times D^{'}$), where each row contains a $DD^{'}$-dimensional adaptive parameter vector,  and 2) the signal matrix $\textbf{H}^{N}_{T} \in \mathbb{R}^{N \times D}$, where each row contains a $D$-dimensional signal vector. Notably, $\textbf{H}^{N}_{T}$ is acquired from the last GCRN layer and can be regarded as a unified latent signal representation of all $T$ time steps.

The core of NMPL is generating node-specific convolutional filter weights from the node adaptive parameters $\Theta^{N} \in \mathbb{R}^{N \times DD^{'}}$ using a linear transformation parameterized by the weight matrix $\textbf{U}\in \mathbb{R}^{DD^{'} \times T L_{F}}$ and convolving the generated filter weights with the unified signal representation $\textbf{H}^{N}_{T}$, where $L_{F}$ is the length of each filter. This process enables gaining three significant benefits: 1) enriching the local node adaptive parameters with node-specific patterns captured globally from the input signals as the weight matrix $\textbf{U}$ is learned during end-to-end training, 2) dynamically synthesizing weights, which is useful in learning a change in a node condition based on a change in the other nodes' conditions since the weights are dynamically adjusted, and 3) attaining multi-task learning across nodes via weight-sharing in the linear transformation matrix $\textbf{U}$. It is worth noting that we do not introduce new model parameters except the weight matrix $\textbf{U}$ since we reuse $\Theta^{N}$ as input to the linear transformation to develop convolutional filters.   
\begin{figure}[h]
	\centering
		\includegraphics[width=1.0\linewidth]{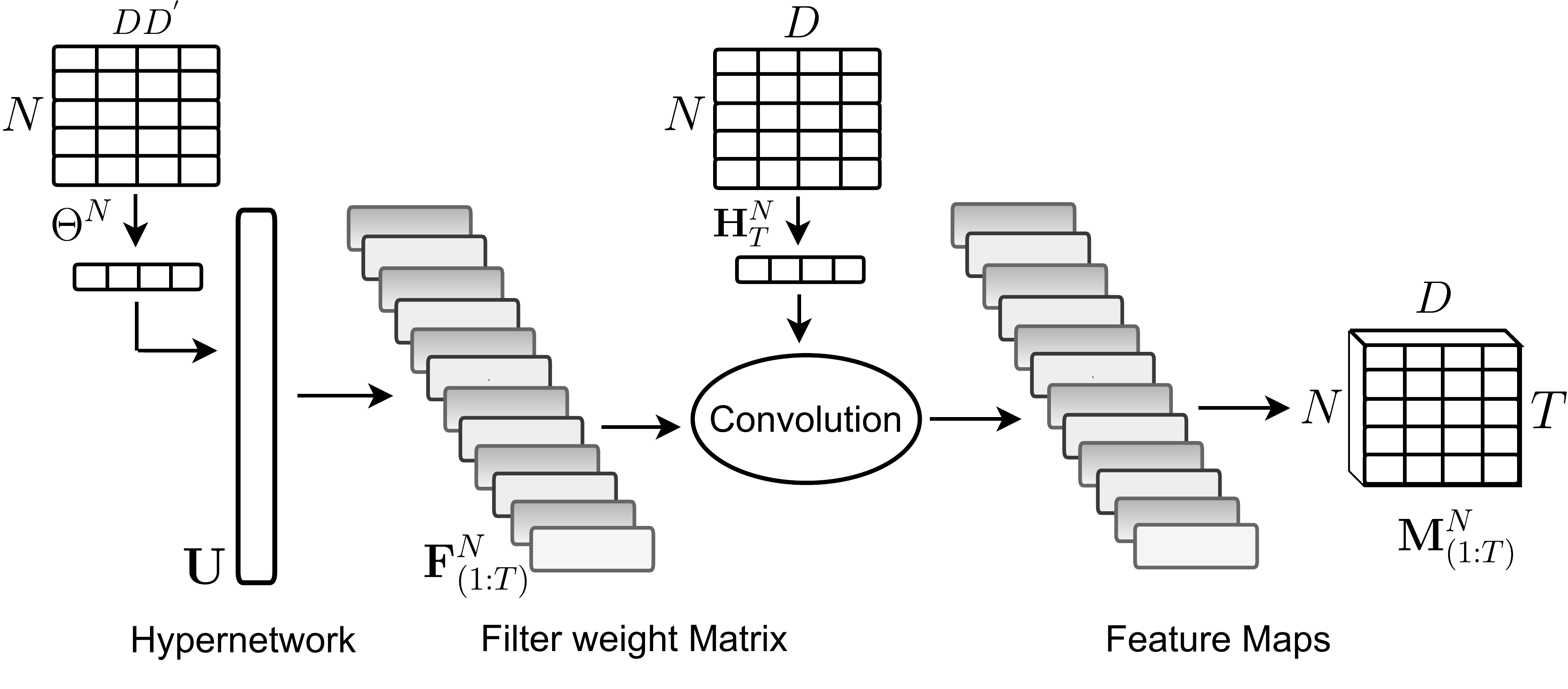}
	\caption{Overall operation of the NMPL module. The hypernetwork $\textbf{U}$ generates $T$ node-specific convolutional filters $\textbf{F}_{(1:T)}^{N}$ from $\Theta^{N}$, each is convolved with $\textbf{H}^{N}_{T}$. The results are obtained in a matrix $\textbf{M}^{N}_{(1:T)}$ comprising $T$ vectorized node-specific feature maps.} 
        \label{fig:fig3}
\end{figure}

In NMPL, we implement the weight matrix $\textbf{U} \in \mathbb{R}^{DD^{'} \times T L_{F}}$ as a hypernetwork. Specifically, a hypernetwork is a meta-learning technique by which one network is used to generate convolutional filter weights for another network. For each node, we keep the number of filters equivalent to the number of output horizons $T$. This implies generating $T$ convolutional filters for each node, each responsible for extracting the corresponding node-specific patterns at each horizon and learning them in the shared latent space $\textbf{U}$. 

Once the inputs are obtained, the hypernetwork is applied to the node adaptive parameter matrix $\Theta^{N} \in \mathbb{R}^{N \times DD^{'}}$, which projects a $DD^{'}$-dimensional weight vector of each node to a $T L_{F}$-dimensional space. The result is reshaped to generate a set of $T$ convolutional filter weights for each node as follows:
\begin{equation}
    \textbf{F}^{N}_{(1:T)} = vec^{-1}(\textbf{U}(\Theta^{N})),
\end{equation}
where $\textbf{F}^{N}_{(1:T)}$ denotes the matrix of node-specific convolutional filters, which comprises a total of $T$ filter weight vectors for each node and operator $vec^{-1}$ denotes vector to matrix reshaping. Although the overall dimensionality of the filter set is set to $T L_{F}$, the rank is limited to $DD^{'}$ to enable parameter sharing between nodes. The signal matrix $\textbf{H}^{N}_{T} \in \mathbb{R}^{N \times D}$ is then convolved with the set of node-specific convolutional filters $\textbf{F}^{N}_{(1:T)}$ as follows:
\begin{subequations}
\begin{align}
    \textbf{M}^{N}_{(1:T)} & = \textbf{H}^{N}_{T} * \textbf{F}^{N}_{(1:T)}, \\
& =   \sum_{n = 1}^{N} \sum_{t = 1}^{T} \textbf{H}^{N}_{T} * \textbf{F}^{n}_{t},
\label{eq:7a}
\end{align}
\end{subequations}
where $*$ represents 1D convolution. The above convolutional operation generates a set of $T$ 1D feature maps $\textbf{M} \in \mathbb{R}^{D}$ for each node $n \in N$, i.e., $\textbf{M}^{N}_{(1:T)} \in \mathbb{R}^{N \times T \times D}$, where $D$ is the length of each node-specific feature map. 

These resultant feature maps are of substantial importance as they contain the specific features of each node extracted during the hypernetwork operation. From a closer view, the NMPL process can be seen as a decoding technique, which is intuitive since GCRNs are primarily designed as pattern encoding techniques and, thus, require an appropriate pattern decoding mechanism before mapping the input signals into the desired scalar outputs.  In more detail, most existing GCRNs define a 1D convolutional layer on the output. The role of this convolutional layer is to obtain the unified signal representation $\textbf{H}^{N}_{T}$  from the last GCRN layer and transform it into scalar predictions of desired horizons. To this effect, a separate filter is defined for each desired horizon that attempts to learn an input-output correlation when mapping the unified signal representation to the corresponding scalar predictions. While maintaining a unique parameter space for each node is essential in the pattern encoding phase, it is also important to extract features of each node and then map those node-specific features to the corresponding predictions rather than directly generating the scalar outputs. NMPL conducts this process as an integral operation and thus gains another significant advantage over many existing multi-step traffic predictors.

To sum up, the NMPL module utilizes a linear transformation weight matrix as a hypernetwork to construct node-specific convolutional filters from the node adaptive parameters, which in turn, extracts the node-specific patterns globally from the signal matrix. These extracted node-specific patterns are used to enrich the node adaptive parameters by learning and updating the hypernetwork parameters during end-to-end training. Meanwhile, the node adaptive parameters become more productive in forming the filter weights since they are generated dynamically from the graph structure. This can better describe the entire weight structure, allow the weights to change across nodes and time steps in the given network, and adapt to the input sequence.

\subsection{Node Attention Weight Generation (NAWG)}
With hypernetwork in the previous module, we capture node-specific patterns so the model learns the unique properties associated with each individual node. Having node-specific features, we can map them to scalar predictions at this stage. (This part is performed as an ablation experiment in Section \ref{sec:ablationstudy}). However,  we observe that NMPL alone is insufficient since it generates the node-specific features from the output of the last GCRN layer,  which tends to struggle in retaining patterns when the sequence grows longer, specifically when the number of nodes is large. Hence, we implement another module NAWG, which receives the full sequence output of the GCRN module $\textbf{H}^{N}_{(1:T)} \in \mathbb{R}^{T \times N \times D}$ as the input and use it to generate attention weights for the node-specific features generated in the NMPL module. Specifically, NAWG is based on the multihead cross-attention mechanism (MHCA) and is served as a bias block for the NMPL module, aiming to generate attention weights for nodes' representations obtained from the NMPL module. It is worth noting that the MHCA is applied to each node individually. 
\begin{figure}[h]
	\centering
		\includegraphics[width=1.0\linewidth]{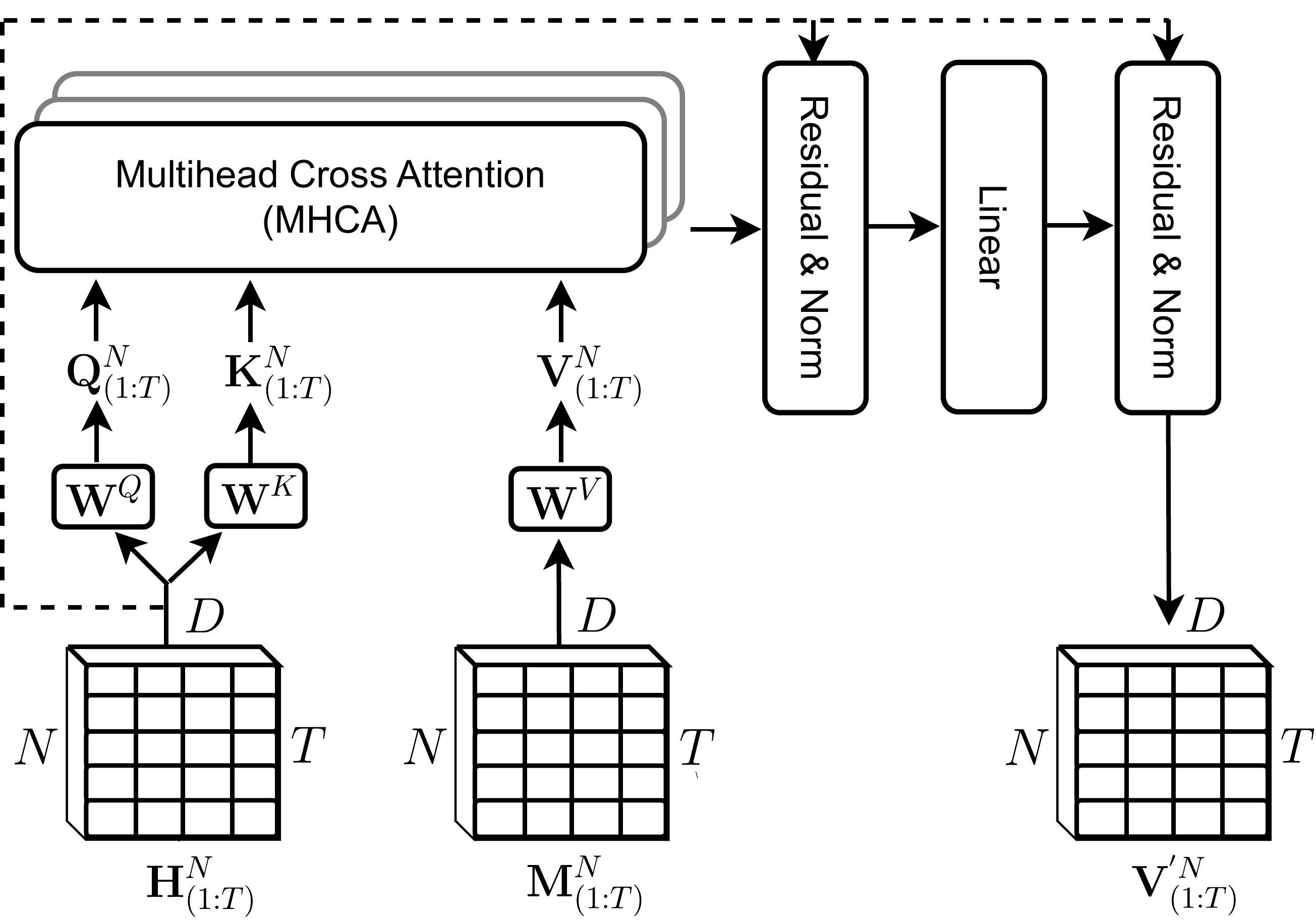}
	\caption{Overall operation of the NAWG module. }
        \label{fig:fig3}
\end{figure}

Let us demonstrate the process of this module via singlehead attention and then describe how it is extended to the multihead cross-attention mechanism. A singlehead attention function receives $D$-dimensional queries $\textbf{Q}$, keys $\textbf{K}$, and values $\textbf{V}$ of all positions in the sequence. The attention weights for each position in the sequence are obtained by computing a dot-product of a given query $\textbf{Q}$ with all keys $\textbf{K}$ followed by a division on the scaling factor $\sqrt{D_{K}}$ and passing the results through a $\text{softmax}$ function. These attention weights are then assigned to the corresponding values $\textbf{V}$, where a weighted sum of them is generated as the output. In practice, this process is conducted for all values simultaneously, which is mathematically represented as follows:
\begin{equation}
    \text{Attention}(\textbf{Q}, \textbf{K}, \textbf{V}) = \text{softmax}\Big(\frac{\textbf{Q}\textbf{K}^{\intercal}}{\sqrt{D_{K}}}\Big) \textbf{V}.
\end{equation}

Existing attention-based traffic predictors mainly utilize the self-attention mechanism \cite{Vaswani2017AttentionIA}, where all queries, keys, and values are generated from the same sequence. We adopt a different approach of cross-attention, where one sequence is used to generate attention weights for another sequence. Specifically, in our case, $\textbf{Q}$ and $\textbf{K}$ for all nodes are obtained from the output of the GCRN module, which we compose in a matrix form $\textbf{H}^{N}_{(1:T)} \in \mathbb{R}^{T \times N \times D}$ by stacking the output of all GCRN layers row-wise in sequential order. Each GCRN layer corresponds to a specific time step in the input sequence. The obtained matrix is then reshaped into $\textbf{H}^{N}_{(1:T)} \in \mathbb{R}^{N \times T \times D}$ since we are computing attention weights for each individual node. $\textbf{H}^{N}_{(1:T)}$ is then projected to the corresponding queries and keys as follows: 
\begin{equation}
    \textbf{Q}^{N}_{(1:T)}  = \textbf{H}^{N}_{(1:T)}  \textbf{W}^{Q}, \textbf{K}^{N}_{(1:T)}  = \textbf{H}^{N}_{(1:T)} \textbf{W}^{K}, 
\end{equation}
where $\textbf{W}^{Q} \in \mathbb{R}^{D \times D}$ and $\textbf{W}^{K} \in \mathbb{R}^{D \times D}$ are the learnarable weight matrices shared by all nodes. Having queries and keys, we compute the attention weights $\textbf{A}^{N}_{(1:T)}$ for all nodes as follows:
\begin{equation}
    \textbf{A}^{N}_{(1:T)} = \text{softmax}\Big(\frac{(\textbf{H}^{N}_{(1:T)}\textbf{W}^{Q}) (\textbf{H}^{N}_{(1:T)}\textbf{W}^{K})}{\sqrt{D_{K}}}\Big).
\end{equation}
Recall Section \ref{sec:sectionNMTM} that the output of the NMPL module is a feature matrix $\textbf{M}^{N}_{(1:T)} \in \mathbb{R}^{N \times T \times D}$. Here, we treat it as a value matrix for which we computed the attention weights from the queries and keys. Hence, for the singlehead cross-attention (SHCA) mechanism, each row of $\textbf{M}^{N}_{(1:T)}$ is first projected to a value, and the corresponding attention weight is assigned to each value to get the output $\textbf{V}^{'N}_{(1:T)} \in \mathbb{R}^{N \times T \times D}$ as follows:
\begin{subequations}
    \begin{align}\textbf{V}^{N}_{(1:T)}  &= \textbf{M}^{N}_{(1:T)} \textbf{W}^{V}, \\
    \textbf{V}^{'N}_{(1:T)}  &= \textbf{A}^{N}_{(1:T)} \textbf{V}^{N}_{(1:T)},
\end{align}
\end{subequations}
where $\textbf{W}^{V} \in \mathbb{R}^{D \times D}$ is the learnable weight matrix shared by all nodes for values. Each attention weight denotes the relationship strength between a value and a query-key pair. Specifically, these attention weights make the generated node-specific features aware of the features learned initially at each GCRN layer for each time step. This allows to enrich the node-specific features with short- and long-range patterns and handles the recurrence limitation of forgetting the previously learned features. A straightforward approach could be utilizing all GCRN layers directly. However, such an approach may lead to a high degree of signal/feature repetition, which potentially leads to over-fitting. Instead, the proposed method provides a more intuitive solution to use those representations to generate attention weights and use them to establish a connection between the generated and initially learned features for retaining the previously learned patterns and also avoid excessive signals/features repetition and thus over-fitting. 

Having introduced the SHCA in NAWG, we move on to the multihead settings, which allows us to jointly attend to information from different representation subspaces to enhance model expressivity. For the MHCA implementation, we split the queries, keys, and values into different representation subspaces and then perform the attention operations in parallel. The results are concatenated to produce the final output. This process is mathematically represented as follows:  
\begin{subequations}
\begin{align}
    \textbf{V}^{'N}_{(1:T)} &= \text{MHCA}(\textbf{H}^{N}_{(1:T)}, \textbf{M}^{N}_{(1:T)}), \\
    &= [\text{head}_1, \hdots, \text{head}_j], \\
   \text{head}_j &= \text{SHCA}(\textbf{A}^{N}_{
{(1:T)}, {j}}, \textbf{V}^{N}_{
{(1:T)}, {j}}), \\
       & = \textbf{A}^{N}_{
{(1:T)}, {j}} \textbf{V}^{N}_{{(1:T)}, {j}},\\ 
\textbf{A}^{N}_{{(1:T)}, {j}} &= \text{softmax}\Big(\frac{\textbf{Q}^{N}_{(1:T),j} \textbf{K}^{N}_{(1:T),j}}{\sqrt{D_{K/j}}}\Big),
\end{align}
\end{subequations}
where $j$ is the number of attention heads.

Generally, the attention functions model the correlation between sequence elements regardless of their distance, as they do not contain any recurrence or convolution operation. This process enables them to capture the global dependencies but compromises them on the local ones inherent in sequential data. To this end, the attention functions typically use positional encoding to insert positional information on the sequence elements. In our implementation, we no longer require positional encoding as the attention input is obtained from the GCRN layers, where recurrence is sufficient to capture the local dependencies. Once the MHCA process is completed, the output is given to a feedforward neural network layer composed of two linear layers with ReLU. Both the MHCA and the feedforward neural network layer are followed by a residual and normalization layer, which adds the previously learned features and conducts batch normalization. By updating Eq. (\ref{eq:eq6}), the overall operation of MACGRN can mathematically be represented as follows: 
\begin{equation}
\begin{split}
     \textbf{V}^{'N}_{(1:T)} = &\, \text{NAWG}(\text{NMPL}(\text{GCRN}  (\hat{\textbf{A}}, [\textbf{X}^{N}_{(1:T)}, \\ &\, \textbf{H}^{N}_{(1-1:T-1)}]; \Theta^{N}; \Omega^{N}; \Phi); \Psi); \Pi),
\end{split}
\end{equation}
where $\textbf{V}^{'N}_{(1:T)} $ represents the output. $\Psi$ and $\Pi$ symbolize the model parameters in NMPL and NAWG modules. The output is then fed to the output layer.

\subsection{Output Layer}
In the output layer, a fully-connected neural network layer $\mathcal{FC}$ is adapted to transform each final $D$-dimensional feature map to the corresponding scalar prediction: 
\begin{equation}
    \textbf{P}^{N}_{(T + 1:T + T)} = \mathcal{FC}(\textbf{V}^{'N}_{(1:T)}),
\end{equation}
where $\textbf{P}^{N}_{(T + 1:T + T)} \in \mathbb{R}^{N \times T \times 1}$ is reshaped to obtain the final prediction matrix $\textbf{P}^{N}_{(T + 1:T + T)} \in \mathbb{R}^{T \times N}$, which is given to the loss function for model training. 

\subsection{Training and Optimization}
We use standard multi-step traffic forecasting settings \cite{Bai2020AdaptiveGC, Wu2019GraphWF, Wu2020ConnectingTD} to train MAGCRN. The historical traffic sequence is split into training, validation, and test sets. Then, a $T + T$ size window is slid over the training set to generate training sequences. The first $T$ elements in each sequence are used as the input, and the last $T$  elements are used as the outputs. Both the inputs and outputs are fed to $\text{L1}$ loss function $\mathcal{L}$:  
\begin{equation}
    \mathcal{L}(\textbf{X}^{N}_{(T + 1:T + T)}, \textbf{P}^{N}_{(T + 1:T + T)}) = \sum_{t = 1}^{T} |\textbf{X}^{N}_{T + t} - \textbf{P}^{N}_{T + t}|,
\end{equation}
where $\textbf{X}^{N}_{(T + 1:T + T)}$ represents the ground truths. The model is trained and optimized with back-propagation and Adam \cite{Kingma2015AdamAM}.
\section{Experiments}
In this section, we first describe the experimental setup, including datasets, evaluation protocols, hyperparameters, implementation details, and baselines. Then, we experiment with traffic flow forecasting and compare the results with baselines to examine the performance gain on both short and long ranges. We also conduct a case study that compares the predictions of MAGCRN against the baselines and ground truths to highlight the prediction quality. In addition, we conduct an ablation experiment to highlight the individual contribution of each module, perform a hyperparameter test to assess model sensitivity, and compare computational costs against the top-performing baselines. Finally, we test MAGCRN on speed forecasting to assess the model generality.   

\subsection{Datasets Details}
To evaluate the performance of the proposed model, we use six real-world traffic datasets: PeMSD3, PeMSD4, PeMSD7, PeMSD8,  PeMSD7(M), and PeMSD7(L). These datasets comprise real-time highway traffic data of California recorded every 30 seconds by the Caltrans Performance Measure System (PeMS). The statistical information of the reported datasets is summarized in TABLE \ref{tab:tab1}. 
\begin{table}[h]
	\caption{Datasets Details} 
	\setlength{\tabcolsep}{1.5 pt}
		\begin{tabular}{l ccccc}
			\toprule
			\multicolumn{1}{l}{Dataset} & \multicolumn{1}{c}{Nodes} & \multicolumn{1}{c} {Time steps} & \multicolumn{1}{c}{Time Interval} & \multicolumn{1}{c} {Type} & \multicolumn{1}{c} {Time Period} \\ 
			\midrule
		      PeMSD3 & 358 & 26,208 & 5 minutes & Flow & 2018/09 - 2018/11\\
                PeMSD4 & 307 & 16,992 & 5 minutes & Flow & 2018/01 - 2018/02\\
                PeMSD7 & 883 & 28,224 & 5 minutes & Flow & 2017/05 - 2017/08\\
                PeMSD8 & 170 & 17,856 & 5 minutes & Flow & 2016/07 - 2016/08\\
                PeMSD7(M) & 228 & 12,672 & 5 minutes & Speed & 2012/05 - 2012/06\\
                PeMSD7(L) & 1026 & 12,672 & 5 minutes & Speed & 2012/05 - 2012/06\\
                \bottomrule
	\end{tabular}
        \label{tab:tab1}
\end{table}

\subsection{Implementation Details}
For all the reported datasets, we follow the standard settings: the raw traffic readings are aggregated into 5-minute interval, making 288 data points per day, 12 steps of recorded data are composed as the input, and the following 12 steps of data is composed as the output, i.e., one-hour's immediate past data is used to predict the next hour's data, the reported datasets are split into training, validation, and test sets with a ratio of 6:2:2, the traffic readings are normalized via standard normalization technique, and the missing values are filled via linear interpolation \cite{Bai2020AdaptiveGC}. The model is trained for 100 epochs with early stop patience of 15 epochs. The performance is evaluated using Root Mean Square Error (RMSE), Mean Absolute Error (MAE), and Mean Absolute Percentage Error (MAPE).
We implement MAGCRN using PyTorch version 1.10.2 and run on NVIDIA GeForce RTX 2080 Ti GPUs with 11 GB memory. The hyperparameters are determined by the model performance on the validation set of each dataset. The number of attention heads is 4, the attention head dimension is 16, and the hidden units ($D$) are 64. The other optimal hyperparameters are provided in TABLE  \ref{tab:tab2}.

\begin{table*}[h]
	\caption{Hyperparameters} 
	\centering
	\setlength{\tabcolsep}{3pt}
		\begin{tabular}{l ccccc}
			\toprule
			\multicolumn{1}{l}{Dataset} & \multicolumn{1}{c}{Embedding dimension ($C$)} & \multicolumn{1}{c}{Filter length $(L_{F}$)} & \multicolumn{1}{c} {Number of attention layers $(L)$} & \multicolumn{1}{c}{Batch size (B)} & \multicolumn{1}{c} {Learning rate (LR)}  \\ 
			\midrule
		      PeMSD3 & 22 & 3 & 1 & 64 & 0.003\\
                PeMSD4 & 8  & 9 & 2 & 64 & 0.003\\
                PeMSD7 & 5  & 3 & 3 & 16 & 0.003\\
                PeMSD8 & 5  & 11 & 1 & 64 & 0.005\\
                PeMSD7(M) & 8  & 3 & 3 & 16 & 0.003\\
                PeMSD7(L) & 8  & 9 & 1 & 16 & 0.003\\
                \bottomrule
	\end{tabular}
        \label{tab:tab2}
\end{table*}
\begin{table*}[h]
	\caption{Flow prediction results} 
	\centering
	\setlength{\tabcolsep}{4pt}
		\begin{tabular}{l ccccccccccccccc}
			\toprule
			\multicolumn{1}{l}{} & \multicolumn{3}{c}{PeMSD3} & & \multicolumn{3}{c} {PeMSD4} & & \multicolumn{3}{c} {PeMSD7} & & \multicolumn{3}{c} {PeMSD8}\\ 
			\cmidrule{2-4}\cmidrule{6-8}\cmidrule{10-12}\cmidrule{14-16}
			 Model & MAE & RMSE & MAPE & & MAE & RMSE & MAPE & & MAE & RMSE & MAPE & & MAE & RMSE & MAPE\\
			\midrule
                STGCN & 17.49 & 30.12 & 17.15 & & 22.70 & 35.55 & 14.59 & & 25.38 & 38.78 & 11.08 & & 18.02 & 27.83 & 11.40  \\
                STSGCN & 17.48 & 29.21 & 16.78 & & 21.19 & 33.65 & 13.90 & & 24.26 & 39.03 & 10.21 & & 17.13 & 26.80 & 10.96 \\
                GWNET & 19.12 & 32.77 & 18.89 & & 24.89 & 39.66 & 17.29 & & 26.39 & 41.50 & 11.97 & & 18.28 & 30.05 & 12.15 \\
                LSGCN & 17.94 & 29.85 & 16.98 & & 21.53 & 33.86 & 13.18 & & 27.31 & 41.46 & 11.98 & & 17.73 & 26.76 &  11.20 \\
                STFGNN & 16.77 & 28.34 & 16.30 & & 19.83 & 31.88 & 13.02 &  & 22.07 & 35.80 & 9.21 & & 16.64 & 26.22 & 10.60 \\
               \addlinespace
              DCRNN & 18.18 & 30.31 & 18.91 & & 24.70 & 38.12 & 17.12 &  & 25.30 & 38.58 & 11.66 & & 17.86 & 27.83 & 11.45 \\
              STGNN & 20.18 & 32.81 & 20.83 & & 23.61 & 37.12 & 16.02 & & 31.69 & 47.84 & 14.08 & & 17.36 & 27.70 & 11.35 \\
              TGCN & 18.93 & 30.91 & 21.59 & & 22.60 & 35.15 & 15.72 & & 27.80 & 41.09 & 14.05 & & 18.36 & 28.00 & 12.67 \\
              AGCRN & 15.98 & 28.33 & 15.29 & & 19.83 & 32.26 & 12.97 & & 21.43 & 35.49 & 9.03 & & 15.95 & 25.22 & 10.09 \\
              Z-GCNETs & 16.64 & 28.15 & 16.39 & & 19.50 & 31.61 & 12.78 & & 21.77 & 35.17 & 9.25 & & 15.76 & 25.11 & 10.01 \\
              \addlinespace
              ASTGCN & 17.77 & 30.03 & 17.03 & & 21.96 & 34.64 & 15.12 & & 25.73 & 39.51 & 12.81 & & 18.29 & 28.09 & 11.56 \\
              MSTGCN & 19.36 & 31.41 & 21.41 & & 24.85 & 37.87 & 19.29 & & 28.31 & 42.75 & 12.88 & & 19.48 & 29.60 & 13.38  \\
              STG2Seq & 19.03 & 29.83 & 21.55 &  & 25.20 & 38.48 & 18.77 & & 32.77 & 47.16 & 20.16 & & 20.17 & 30.71 & 17.32 \\
              DSANet & 21.29 & 34.55 & 23.21 & & 22.79 & 35.75 & 16.03 & & 31.36 & 49.11 & 14.43 & & 17.14 & 26.96 & 11.32 \\
              DSTAGNN & \underline{15.57} & \underline{27.21} & \underline{14.68} & & \underline{19.30} & \underline{31.46} & \underline{12.70} & & \underline{21.42} & \underline{34.51} & \underline{9.01} & & \underline{15.67} & \textbf{24.77} & \underline{9.94} \\
              \addlinespace
              STGODE & 16.50 & 27.84 & 16.69 &  & 20.84 & 32.82 & 13.77 & & 22.99 & 37.54 & 10.14 & & 16.81 & 25.97 & 10.62 \\ 
                \midrule
             MAGCRN (Ours) & \textbf{15.10} & \textbf{26.28} & \textbf{14.08} & & \textbf{19.04} & \textbf{31.20} & \textbf{12.45} & & \textbf{20.47} & \textbf{34.27} & \textbf{8.67} & & \textbf{15.46} & \underline{25.03} & \textbf{9.79} \\
                \bottomrule
	\end{tabular}
        \label{tab:tab3}
\end{table*}

\subsection{Baselines}
For multi-step traffic forecasting, we compare the performance of the proposed MAGCRN against well-known recent state-of-the-art baselines. We categorize these baselines into convolution-based models (STGCN, STSGCN, GWNET, LSGCN, STFGNN), recurrence-based models (DCRNN, STGNN, TGCN, AGCRN, Z-GCNETs), and attention-based models (DSTAGNN, ASTGCN, MSTGCN, STG2Seq, DSANet). These models have been thoroughly discussed in Section \ref{sec:secrelatedwork}. TGCN and STGNN are implemented with multi-step settings \cite{Bai2020AdaptiveGC, chen2021z} to generate predictions as a whole. In addition, we include a recent ordinary differential equation (ODE) based model STGODE in the baselines, which replaces graph convolution with ODE. We exclude the traditional time-series analysis methods, such as ARIMA, VAR, and SVR, from the baseline list since their predictive performance largely lags behind the reported deep models. 

\begin{figure*}[h]
	\centering
        \includegraphics[width=0.99\linewidth]{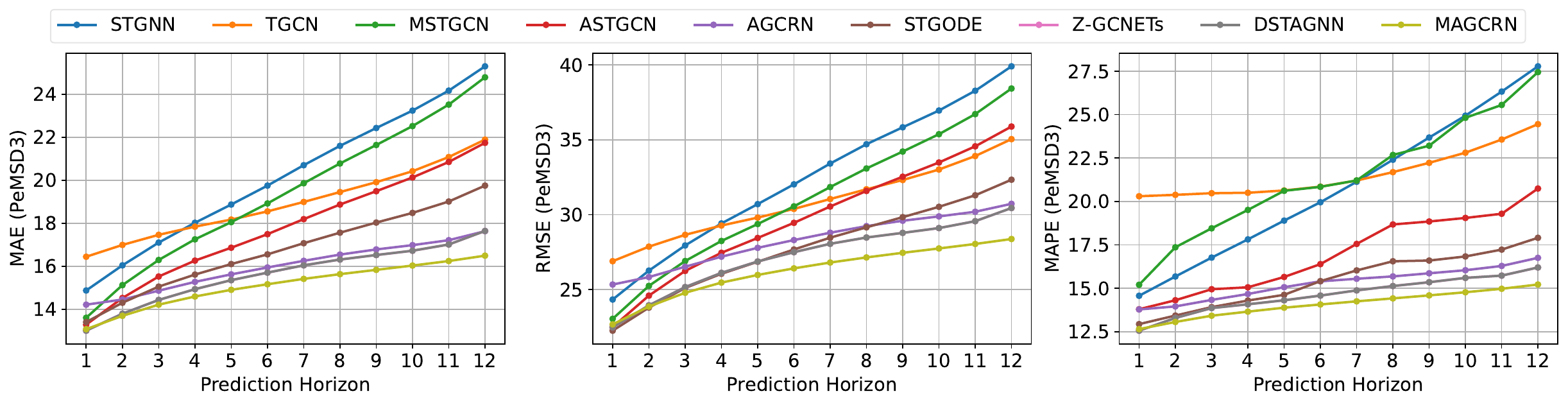} 
        \centering
        \includegraphics[width=0.99\linewidth]{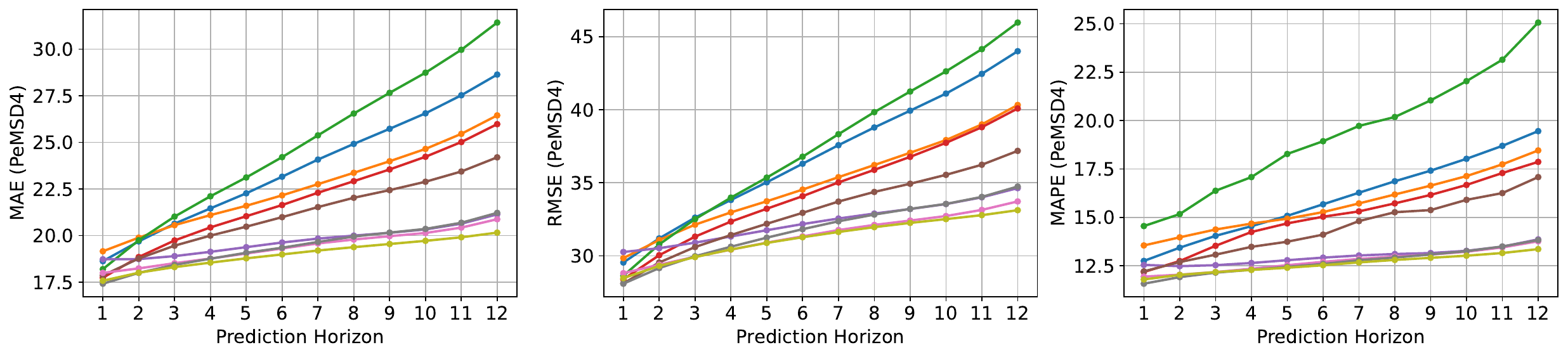}
        \centering
        \includegraphics[width=0.99\linewidth]{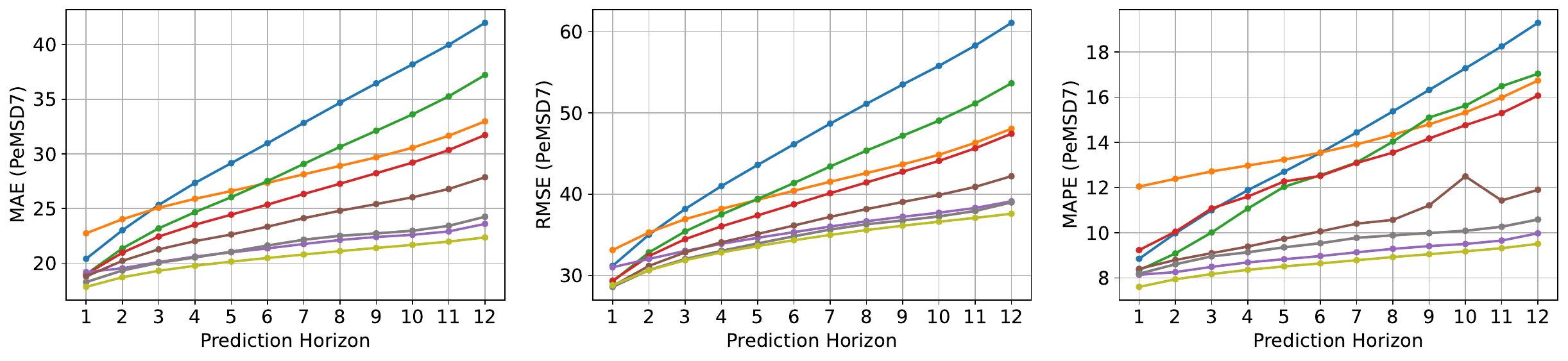} 
        \centering
        \includegraphics[width=0.99\linewidth]{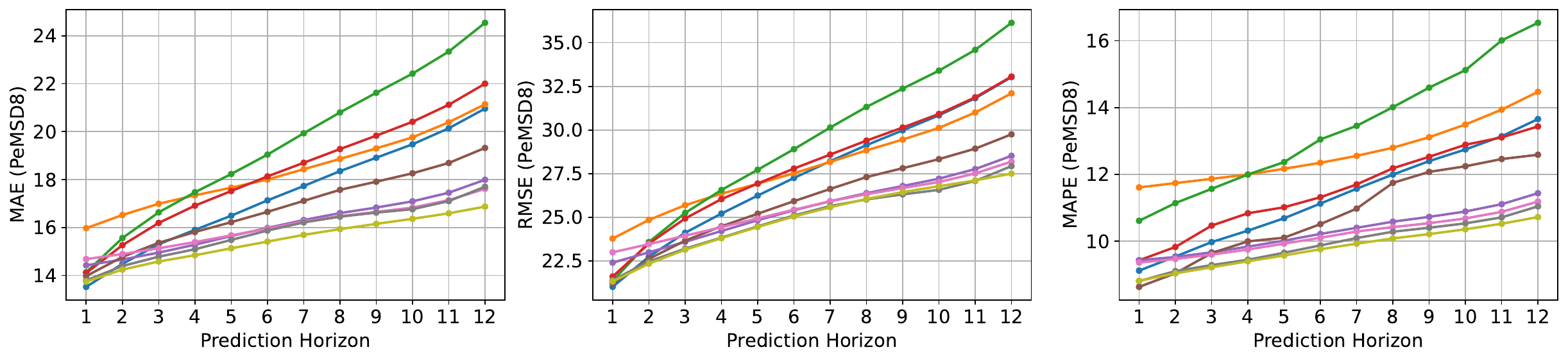}
	\caption{Predictive performance comparison at each horizon. The results of the baselines are reproduced via their publically available source codes for comparison at each prediction interval. }
 \label{fig:fig-pred-pems4-5}
\end{figure*}


\subsection{Results on Flow Prediction} 
TABLE  \ref{tab:tab3} reports the average traffic flow prediction results of the proposed MAGCRN and compares them against the published results of the baselines on PeMSD3,  PeMSD4, PeMSD7, and PeMSD8 datasets. The best scores are highlighted in bold, and the second-best scores are underlined. Except for RMSE on PeMSD8, MAGCRN outperforms all state-of-the-art baselines on each reported metric on all datasets. The performance gap between MAGCRN and the baselines is larger on datasets with a large number of nodes (i.e., PeMSD3 and PeMSD7) than on those with a small number of nodes (i.e., PeMSD4 and PeMSD8). We attribute this performance gain to learning the node-specific patterns and weighing them with initial features, which play a vital role in performance improvement when the number of nodes is large. Considering the rapid growth of traffic networks with urbanization, we believe this observation is of significant importance to future applications.
\begin{figure*}[h]
	\centering
        \includegraphics[width=0.78\linewidth]{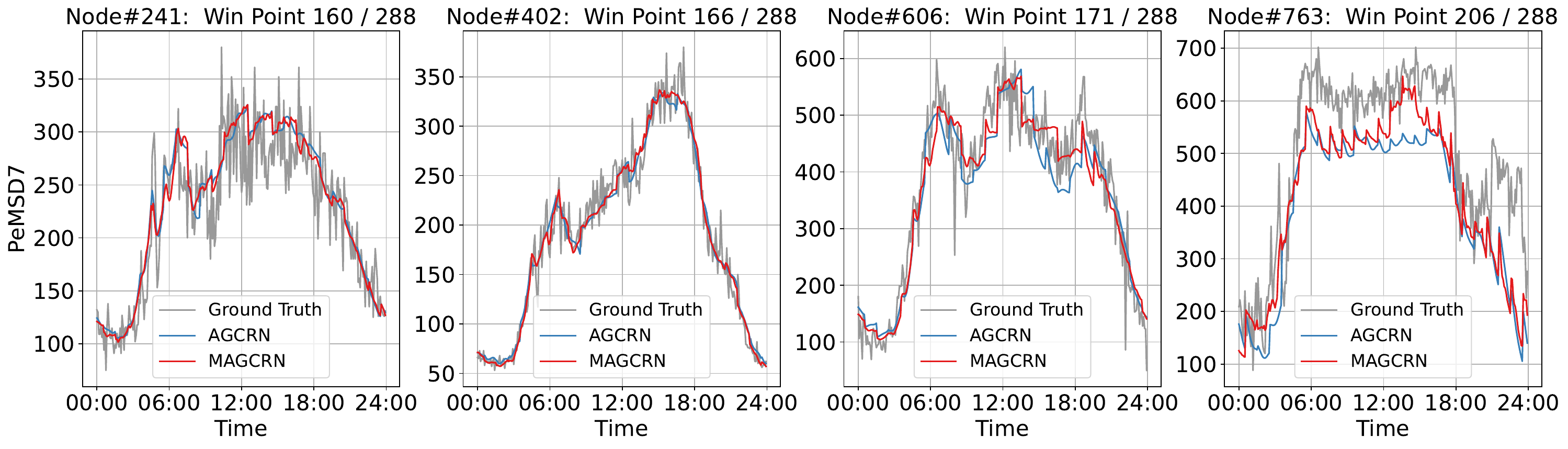} 
        \centering
        \includegraphics[width=0.78\linewidth]{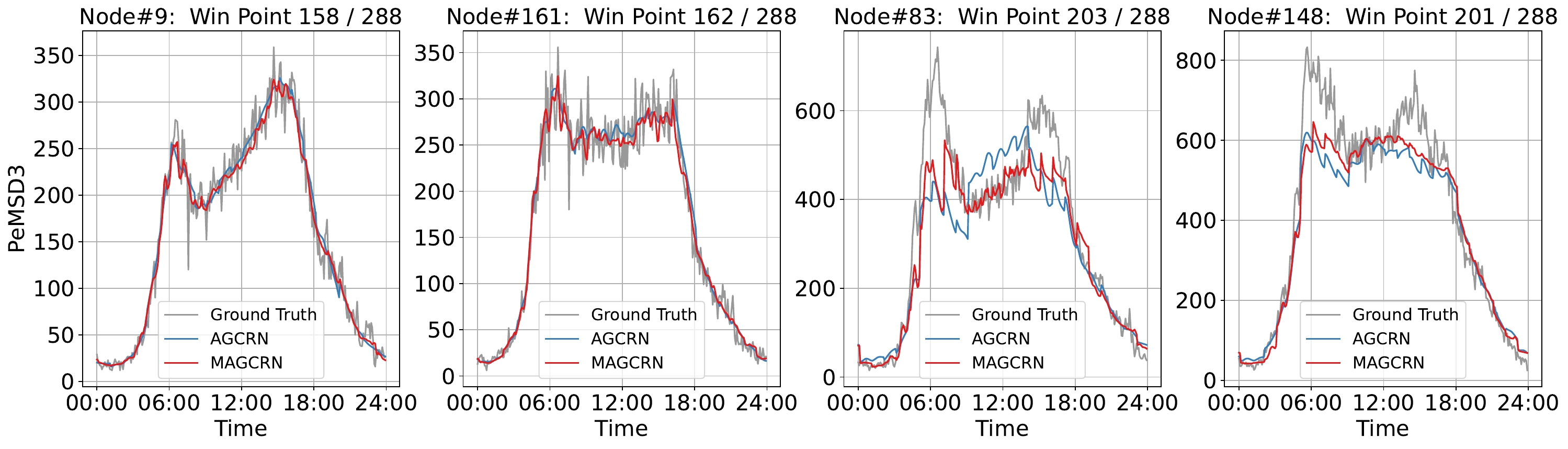}
	\caption{Predictions vs. Ground Truths.}
 \label{fig:fig-pred-pems4}
\end{figure*}
\begin{figure*}[h]
        \centering
        \includegraphics[width=0.78\linewidth]{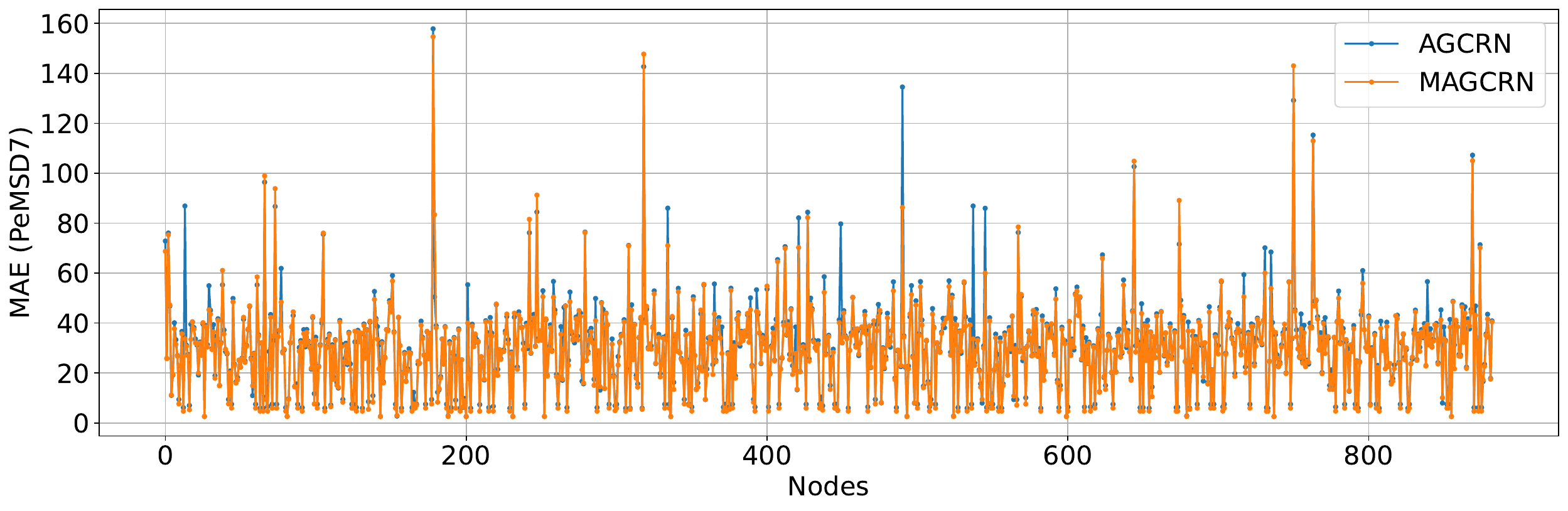}
        \centering
        \includegraphics[width=0.78\linewidth]{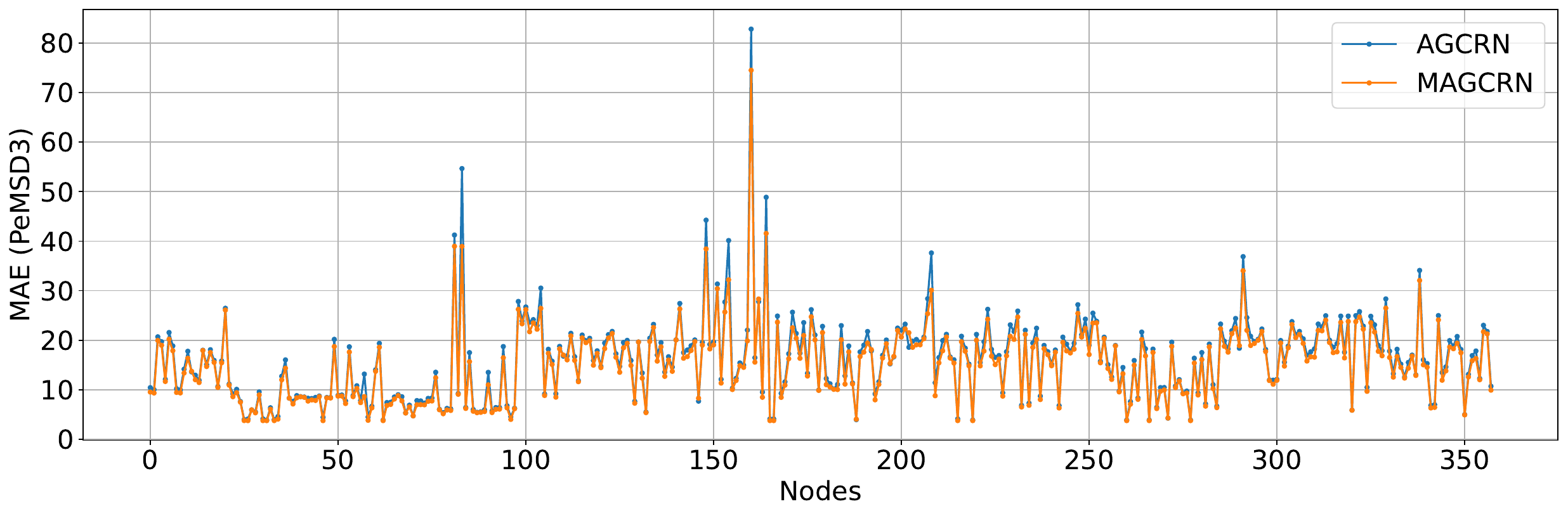}
	\caption{Node-to-node MAE comparison.}
 \label{fig:fig-rmse-pems4}
\end{figure*}

Besides, Fig. \ref{fig:fig-pred-pems4-5} compares the results of MAGCRN with eight baselines on all prediction horizons, among which four are the top-performing baselines in the table, while two recurrence-based and two attention-based methods are additionally included due to their relevancy to the proposed MAGCRN. Generally, the predictive performance of traffic predictors decreases as the prediction interval increases since the task becomes more difficult for large intervals. However, it can be observed that the performance of MAGCRN drops considerably lower than all the baselines. In fact, MAGCRN outperforms the baselines with a larger margin on large intervals than the small ones, particularly on large datasets. For example, on PeMSD4, MAGCRN outperforms the top-performing baseline DSTAGNN with  0.6$\%$, 1.9$\%$, and 5.0$\%$ MAE on the prediction intervals 3, 6, and 12, respectively. In contrast, on the PeMSD7 dataset, MAGCRN outperforms DSTAGNN with 3.6$\%$, 5.3$\%$, and 7.8$\%$ MAE on the prediction intervals 3, 6, and 12, respectively.  

The performance gap is even larger when comparing the results of PeMSD7 with PeMSD8. MACGRN gains an average of 4.43$\%$ MAE improvement on PeMSD7 compared to 1.34$\%$ on PeMSD8 over DSTAGNN since there is a large difference between the number of nodes in both datasets. This comparison reveals a significant benefit of MAGCRN over the baselines for long-term forecasts with a large number of nodes as our model's performance increases with an increase in the prediction interval and more on the larger datasets. This observation supports our notion of generating weights from features learned initially at each GCRN layer. These weights provide a foundation for the node-specific features and thus enhance the predictive performance at each horizon but more on the large intervals since the baselines struggle with the difficult task.  

STGCN, STSGCN, Graph WaveNet, LSGCN, and STFGNN are convolution-based representative methods that utilize 1D or temporal convolution to capture temporal dependencies. 1D convolution is typically limited by the size of the receptive field, making it hard to attend to long-range temporal information. Although temporal convolution handles this limitation with dilation to allow an exponentially large kernel size, it still requires a stack of multiple convolutional layers to connect a pair of positions in a given sequence. It thus makes it harder to make accurate long-term predictions. Attention-based methods such as ASTGCN and DSANet have the advantage of a global receptive field but are limited by their ability to capture local patterns inherent in continuous data and thus fail to obtain a good performance on all prediction intervals. 

On the other hand, DCRNN, TGCN, STGNN, AGCRN, and Z-GCNETs are representative recurrence-based methods, which are limited by the lack of the ability of the recurrence to maintain long-range temporal patterns. However, AGCRN has still shown the best performance and stands amongst the top three baselines in the table, mainly due to learning the node-adaptive parameters. In summary, the proposed MAGCRN largely outperforms all reported baselines since it handles their limitations by effectively integrating recurrence, convolution, and attention to generate the unified representation, learn node-specific patterns, and weigh them with the initially learned features.

\subsection{Case Study}
In this section, we compare the predictions of MAGCRN against the closely related model AGCRN. The goal of this comparison is to illustrate that MAGCRN improves predictive performance by predicting traffic flow values closer to the true ones. We select four nodes from the large datasets PeMSD3 and PeMSD7, generate their one-day-ahead flow predictions, and then compare them against the ground truths in Fig. \ref{fig:fig-pred-pems4}. We use a metric called Win Point, which computes the number of optimal predictions in the prediction matrix. Optimal predictions are referred to as those which are the closest to the ground truths. 

On PeMSD7, nodes 241 and 402 fluctuate approximately to 350. Here, AGCRN  makes 128 and 122 optimal predictions, whereas MAGCRN makes 160 and 166, with a Win percentage of 20$\%$ and 26$\%$, respectively. On the other hand, nodes 606 and 763 fluctuate to a higher level of 600 and 700, respectively. Here, AGCRN  makes 117 and 82 optimal predictions, whereas MAGCRN makes 171 and 206, with a Win percentage of 31$\%$ and 60$\%$, respectively. A similar performance gain can also be observed in the PeMSD3 dataset, which validates the prediction quality of the proposed MAGCRN. In addition, Fig. \ref{fig:fig-rmse-pems4} visualizes a node-to-node MAE comparison of MARGCN and AGCRN on both datasets. The blue dots corresponding to AGCRN are clearly visible across most nodes in both datasets, implying that MAGCRN decreases error on most of the nodes, which is a significant advantage of the MAGCRN.  
 
 \begin{table*}[h]
	\caption{Ablation study} 
	\centering
	\setlength{\tabcolsep}{4pt}
		\begin{tabular}{c ccccccccc}
			\toprule
			 Dataset & & Metric & MAGCRN (w/o NAWG) & MAGCRN (w/o NMPL) & MAGCRN (Query) & MAGCRN (Key) & Default \\
			\midrule
                 & & MAE & 15.46 & 15.32 & \underline{15.21} & 15.43 & \textbf{15.10} \\ 
                 PeMSD3 & & MAPE & 14.68 & 14.58 & 14.58 & \underline{14.27} & \textbf{14.08}  \\
                 & & RMSE & 27.29 & 27.35 & \underline{26.76} & 26.87 & \textbf{26.28}  \\
                 \addlinespace
                 & & MAE & 19.22 & 19.29 & \underline{19.08} & 19.48 & \textbf{19.04} \\ 
                 PeMSD4 & & MAPE & \underline{12.73} & 12.75 & 12.81 & 12.98 & \textbf{12.45}  \\
                 & & RMSE & \underline{31.28} & 31.75 & \underline{31.28} & 31.91 & \textbf{31.20}  \\
                \bottomrule
	\end{tabular}  
        \label{tab:tab5}
\end{table*}
\begin{figure*}[h]
	\centering
        \includegraphics[width=0.80\linewidth]{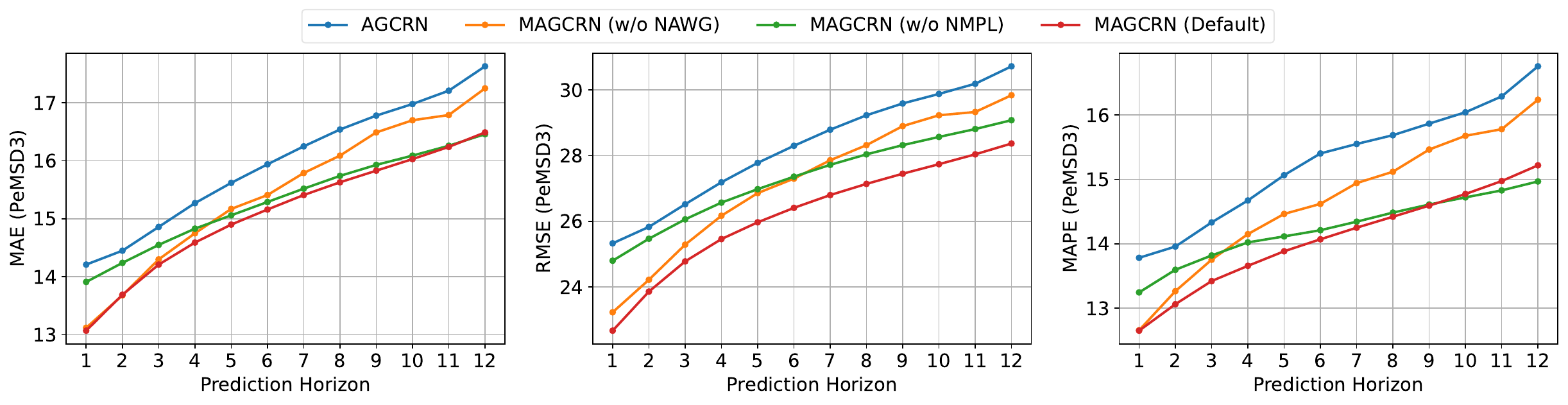} 
        \centering
        \includegraphics[width=0.80\linewidth]{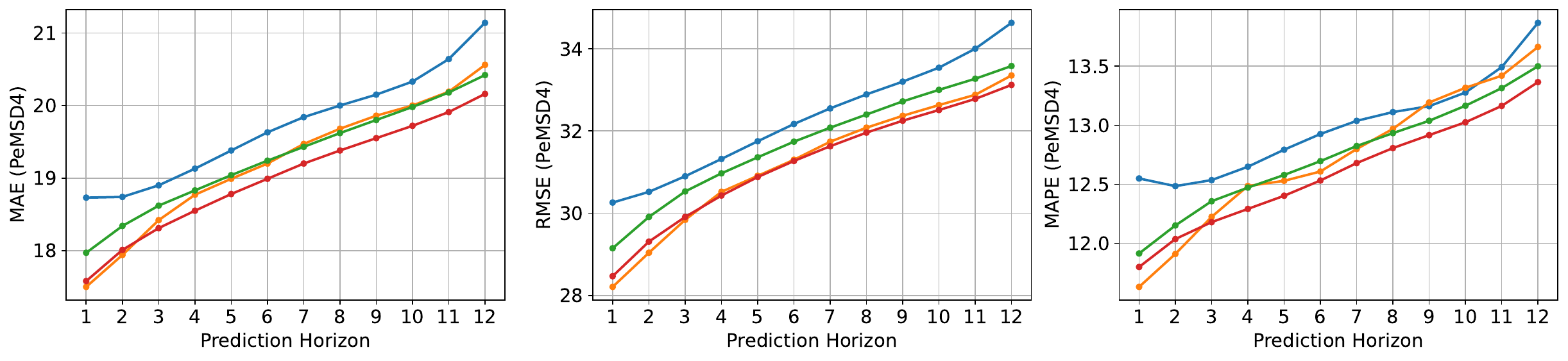}
	\caption{Predictive performance comparison of ablated models at each horizon. }
 \label{fig:fig-pred-pems4-6}
\end{figure*}

\subsection{Ablation Study}
\label{sec:ablationstudy}

TABLE  \ref{tab:tab5} reports the results of the ablation tests performed on PeMSD3 and PeMSD4 datasets. The goal of this ablation experiment is to investigate the contribution of each proposed module to the overall predictive performance. We design the following ablated variants:
\begin{itemize}
    \item MAGCRN (w/o NAWG): It removes the NAWG module to study the specific benefits of the NMPL module.  
   \item MAGCRN (w/o NMPL): It removes the NMPL module and employs the NAWG module directly on the output of the GCRN module. 
   \item MAGCRN (Query): It employs NMPL to generate queries rather than values. 
   \item MAGCRN (Key): It employs NMPL to generate keys rather than values. 
\end{itemize}
All the ablated variants are trained and tested with the same hyperparameter set. Both MAGCRN (w/o NAWG) and MAGCRN (w/o NMPL) cause a considerable drop against the default settings, validating that both learning node-specific patterns and weighing them with initial features contribute to the overall performance. In addition, it can be observed that their contribution varies across both datasets, suggesting that both settings are vital for traffic networks where little is known about the complex nature of traffic data. Besides, the considerable drops in the performance of MAGCRN (Query) and MAGCRN (Key) demonstrate the usefulness of generating values using the NMPL module. 

In addition, Fig. \ref{fig:fig-pred-pems4-6} compares the results of the ablated models MAGCRN (w/o NAWG) and MAGCRN (w/o NMPL) against the default settings on all prediction horizons. AGCRN is included in the comparison as a reference to an individual GCRN. Except for RMSE on PeMSD4, it can be observed that MAGCRN (w/o NAWG) performs better on short intervals and MAGCRN (w/o NMPL) on long intervals, while the default MAGCRN on both short and long intervals, which validates the effectiveness of both NMPL and NAWG modules in handling the recurrence limitations.  
\begin{figure*}
     \centering
     \begin{subfigure}
         \centering
         \includegraphics[width=0.24\linewidth]{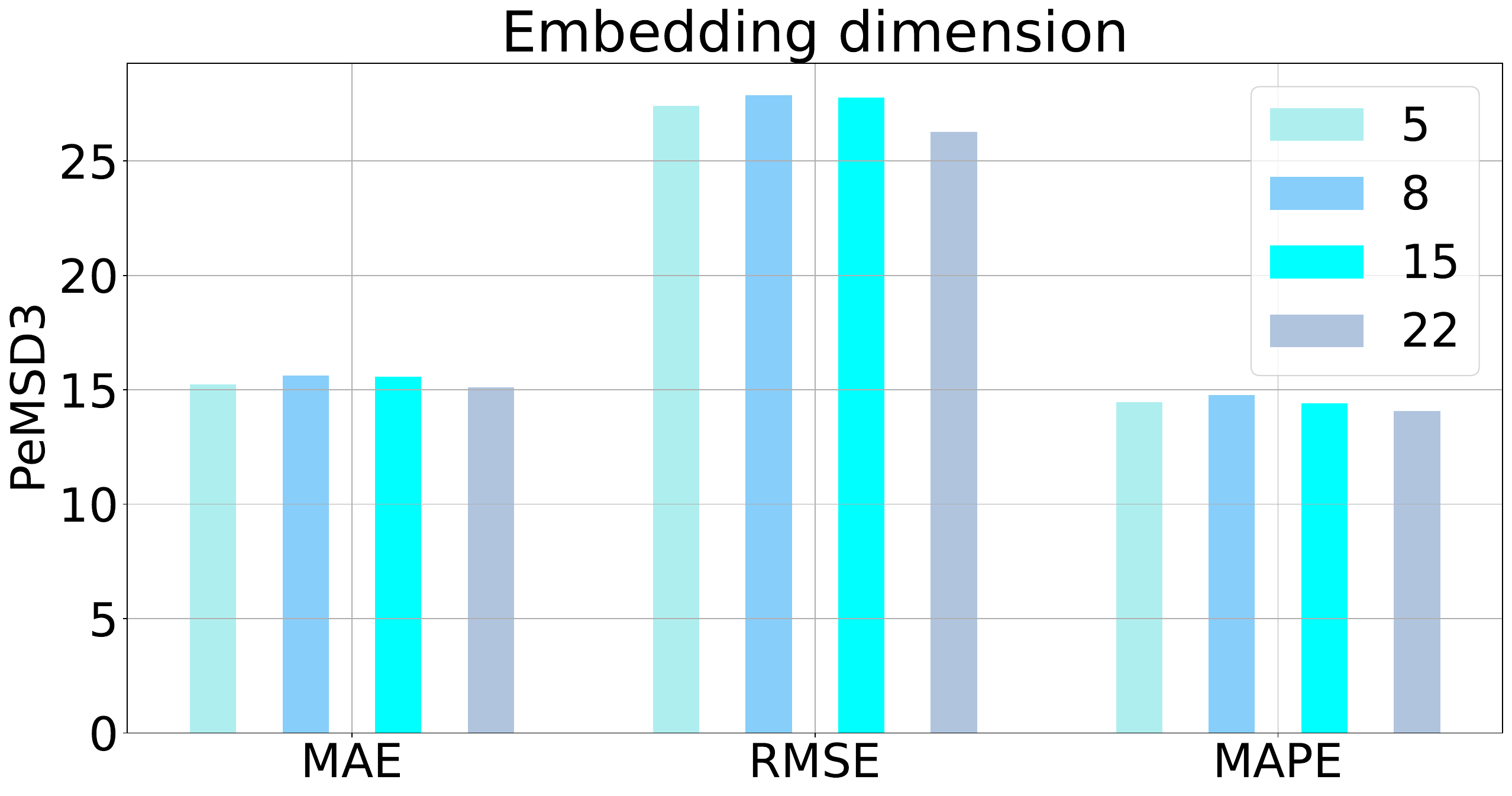}
     \end{subfigure}
     \begin{subfigure}
         \centering
         \includegraphics[width=0.24\linewidth]{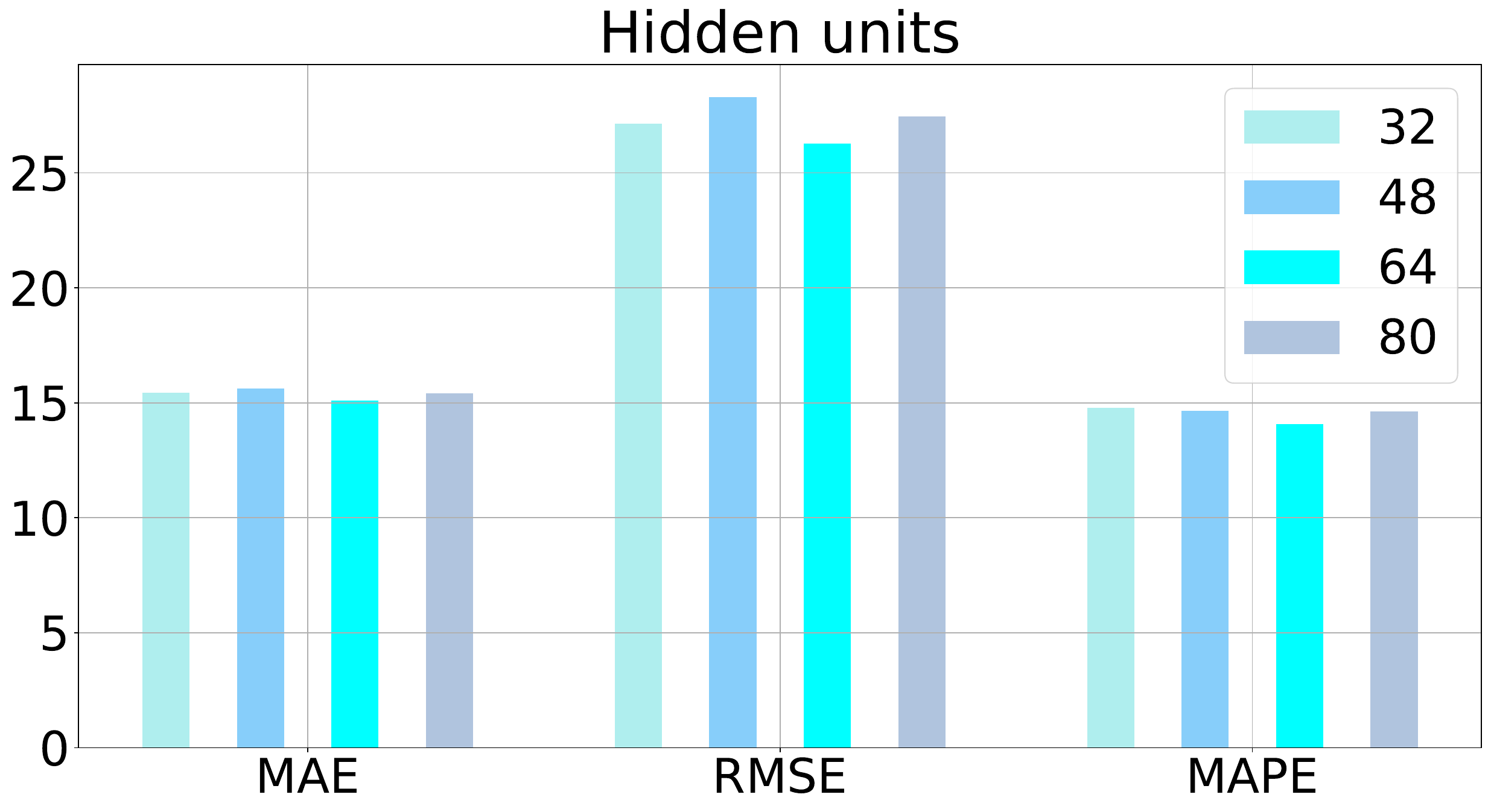}
     \end{subfigure}
     \begin{subfigure}
         \centering
         \includegraphics[width=0.24\linewidth]{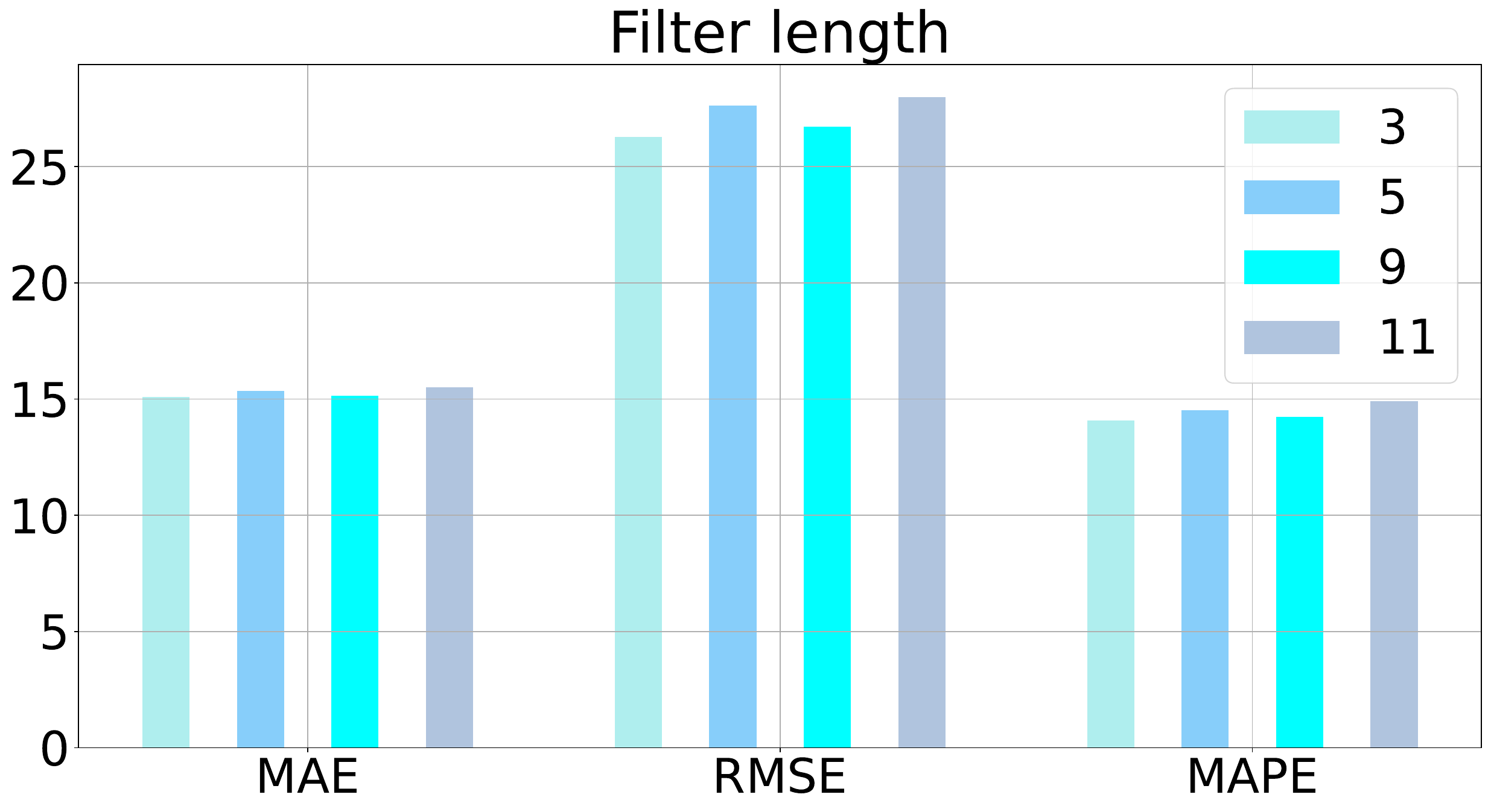}
     \end{subfigure}
     \begin{subfigure}
         \centering
         \includegraphics[width=0.24\linewidth]{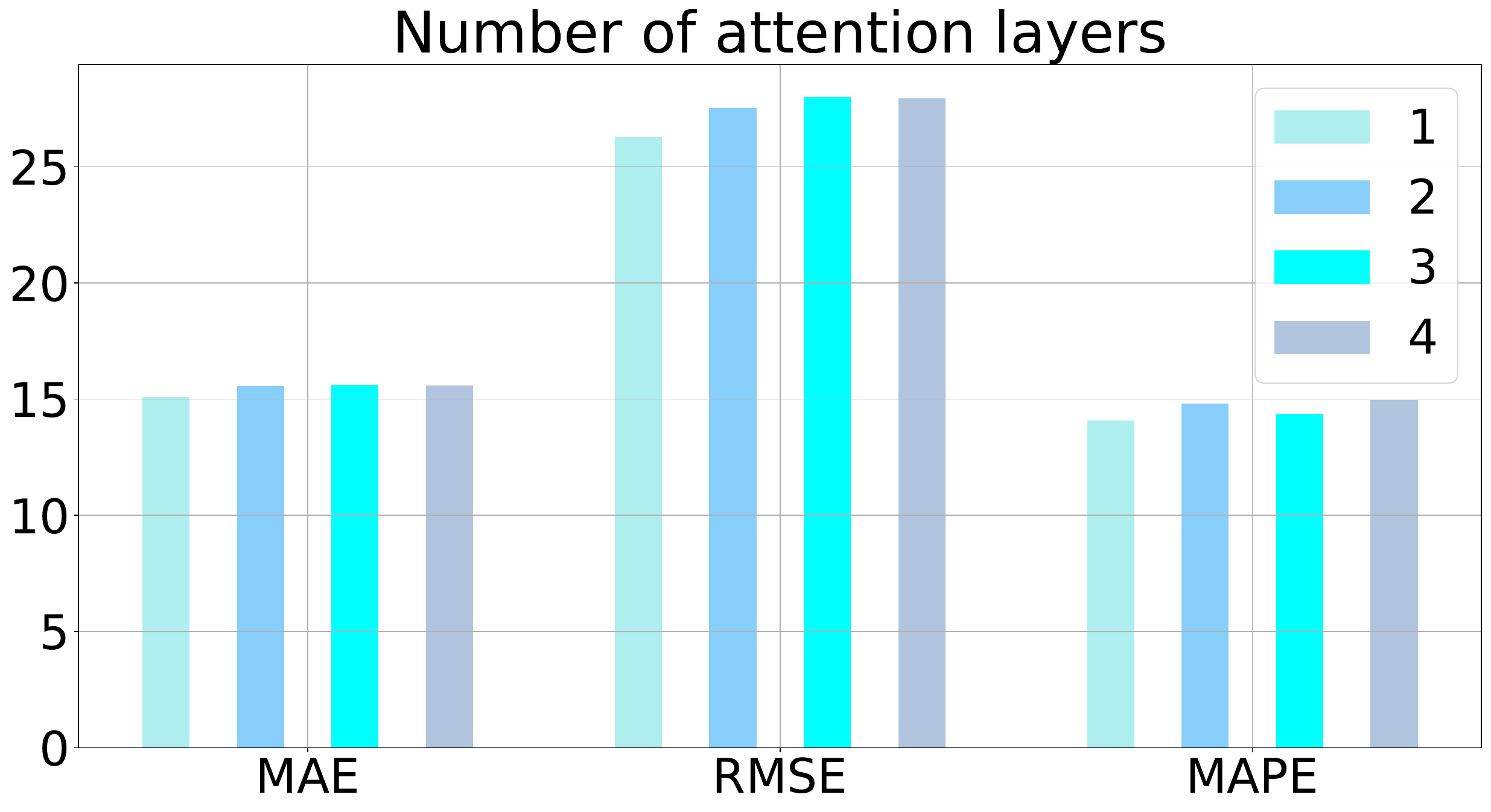}
     \end{subfigure}
     \hfill
     \begin{subfigure}
         \centering
         \includegraphics[width=0.24\linewidth]{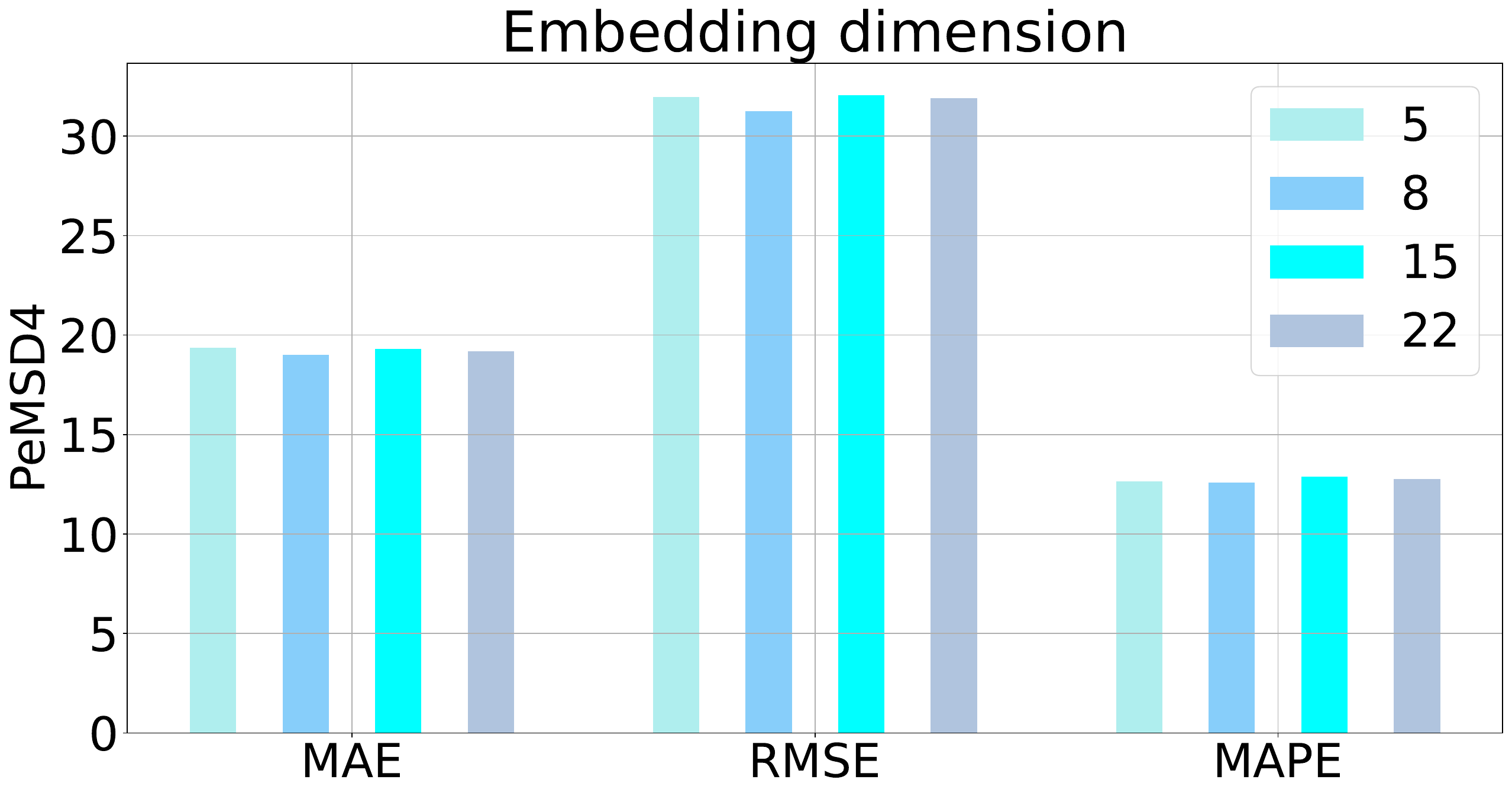}
     \end{subfigure}
     \begin{subfigure}
         \centering
         \includegraphics[width=0.24\linewidth]{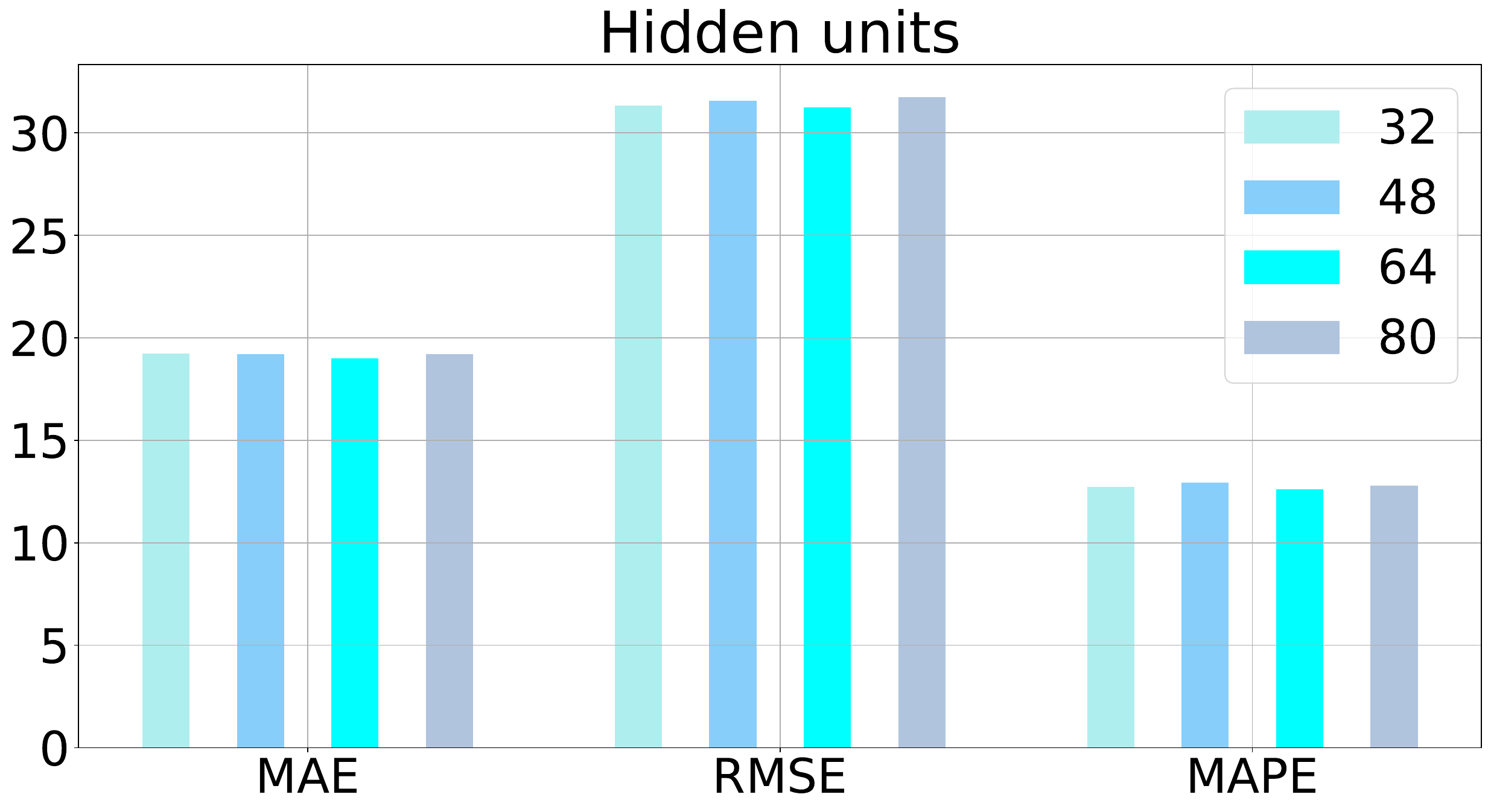}
     \end{subfigure}
     \begin{subfigure}
         \centering
         \includegraphics[width=0.24\linewidth]{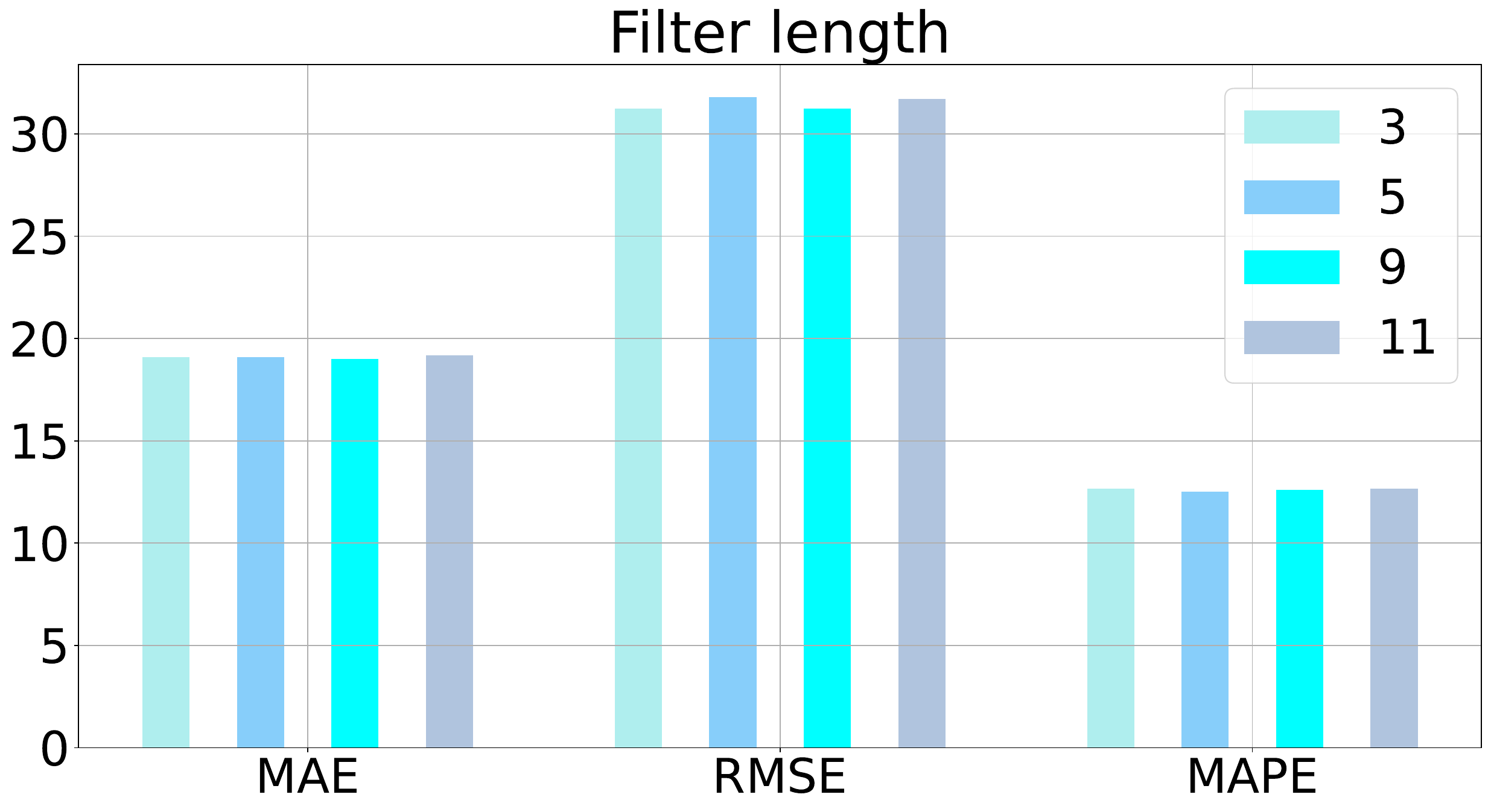}
     \end{subfigure}
     \begin{subfigure}
         \centering
         \includegraphics[width=0.24\linewidth]{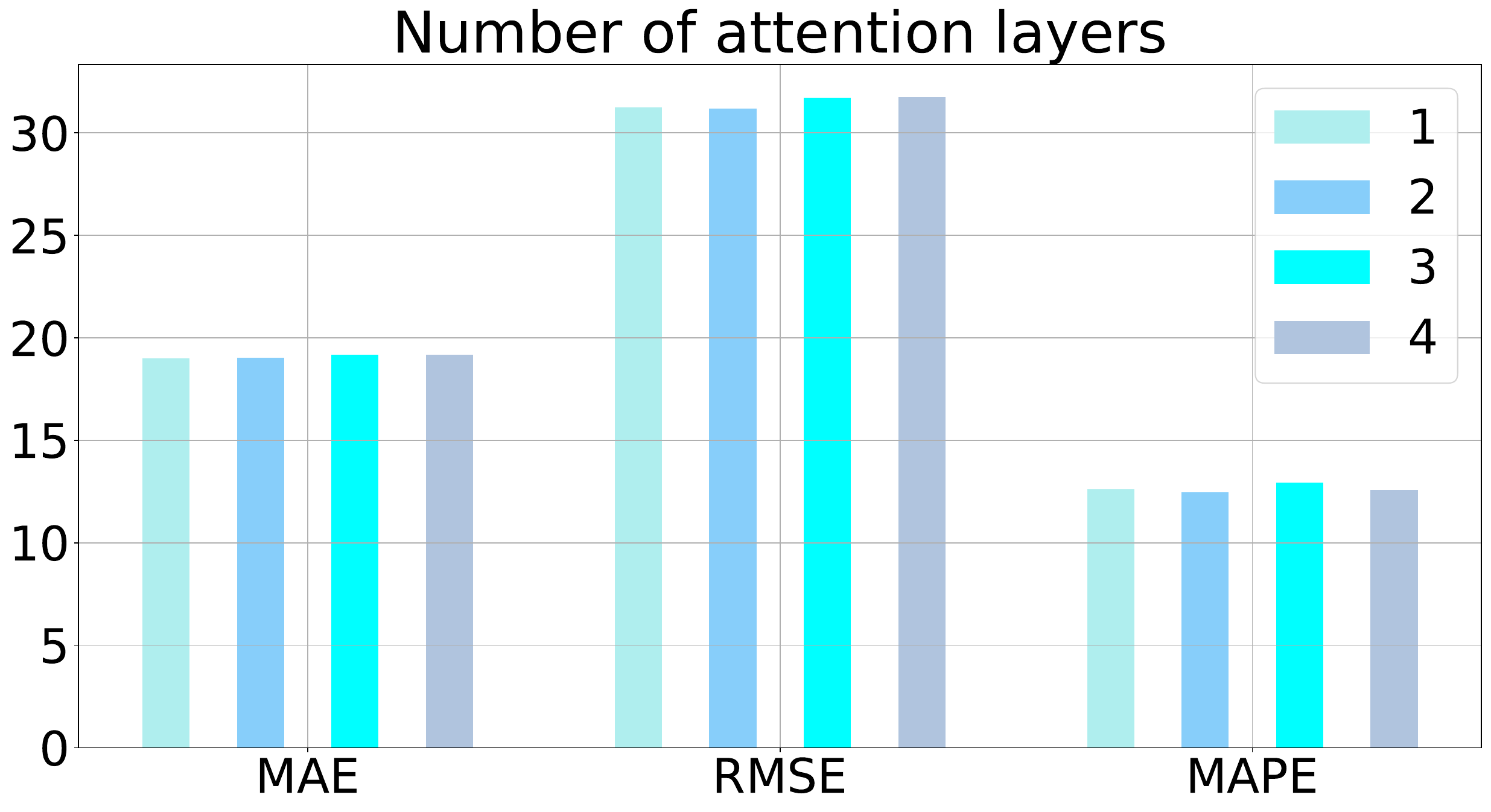}
     \end{subfigure}
        \caption{Effect of different network configuration on PeMSD3 and PeMSD4 datastes.}
        \label{fig:fig9}
\end{figure*}
\subsection{Network Configuration Study}
In this section, we investigate the influence of the hyperparameter settings on the performance of MAGCCN. In particular, we experiment with different hyperparameter configurations on PeMSD3 and PeMSD4 datasets and visualize the MAE, RMSE, and MAPE results in Fig. \ref{fig:fig9}.  In this experiment, we select the most influential hyperparameters: embedding dimension ($C$), hidden units ($D$), filter length ($L_{F}$), and the number of attention layers ($L$), while excluding the ones with limited effect on these datasets. The size of embedding dimensions varies between 5 and 22, while the ones with higher impact are reported. The number of hidden units ranges from 36 to 80, with a gap of 16. The filter length ranges from 3 to 11 with a gap of 2, while the result of length 7 is excluded due to its negligible effect. The number of attention layers is set between 1 and 4, increasing by 1.    

From Fig. \ref{fig:fig9}, it can be observed that MAGCRN is not highly sensitive to hyperparameters, and its response to network configurations differs on distinct datasets. For example, MAGCRN requires a small embedding size of 8, a large kernel size of 9, and two attention layers on PeMSD4, while a large embedding size of 22, a small kernel size of 3, and a single attention layer on PeMSD3 to produce optimal results. This observation, also evident from the results and configurations of many existing methods, suggests that an optimal hyperparameter set is vital for traffic predictors mainly due to the complex nature of data across different parts of the traffic network. Hence, we can conclude that the performance of traffic predictors is not linearly dependent on the values of hyperparameters, and increasing/decreasing the values is not a universal rule in the traffic forecasting domain, but a thorough search is required across a large space on each dataset. 
\begin{table}[h]
	\caption{Computation cost on PeMSD4} 
	\centering
	\setlength{\tabcolsep}{1pt}
		\begin{tabular}{l cc}
			\toprule
			\multicolumn{1}{l}{Model} & \multicolumn{1}{c}{Number of parameters} & \multicolumn{1}{c} {Training time (epoch)} \\ 
			\midrule
                DCRNN & 149.1 K & 28.52 s\\
                TGCN & 13.45 K & 8.22 s\\
                STGNN & 618.76 K & 33.47 s\\
                AGCRN & 748.81 K & 27.71 s\\
                Z-GCNETs & 455.03 K & 41.77 s\\
                ASTGCN  & 450.03 K & 31.42 s\\
		      STGODE & 714.5 K &  51.1 s\\
                DSTAGNN  & 3.6 M & 122.58 s \\
                MAGCRN (w/o NMPL) & 619.54 K & 28.62 s \\
                MAGCRN (w/o NAWG) & 2.37 M & 25.57 s \\
                MAGCRN & 2.39 M & 30.89 s \\
                \bottomrule
	\end{tabular}
        \label{tab:tab6}
\end{table}
\subsection{Computation Cost}
To assess feasibility, we compare the training time and parameter number of MAGCRN with the baselines on the PeMSD4 dataset in TABLE \ref{tab:tab6}. TGCN requires both a small number of parameters and a short training time. However, its ability to learn spatial-temporal correlations is weaker and can not produce state-of-the-art performance. MAGCRN requires almost three times more parameters than the third-best baseline AGCRN due to implementing two additional modules for learning more refined features. Regarding training time, MAGCRN runs approximately three seconds per epoch slower than the AGCRN. However, such a slight delay in training is negligible since the overall performance improvement over AGCRN is highly notable. Besides, removing NMPL from MAGCRN reduces the parameters by approximately 3.86 times, which brings the parameter number lower than AGCRN since we do not use the 1D convolutional layer on top of the GCRN module. On the other hand, removing NAWG does not much affect the parameter number but reduces the training time by 5 seconds per epoch. Z-GCNETs require a smaller number of parameters but a longer running time than MAGCRN. Compared with the top-performing baseline DSTAGNN, MAGCRN is highly efficient in terms of both model parameters and training time. In summary, MAGCRN has moderate computation cost while having the best predictive performance compared to the recent state-of-the-art traffic predictors. 

\begin{figure}[h]
	\centering
		\includegraphics[width=\linewidth]{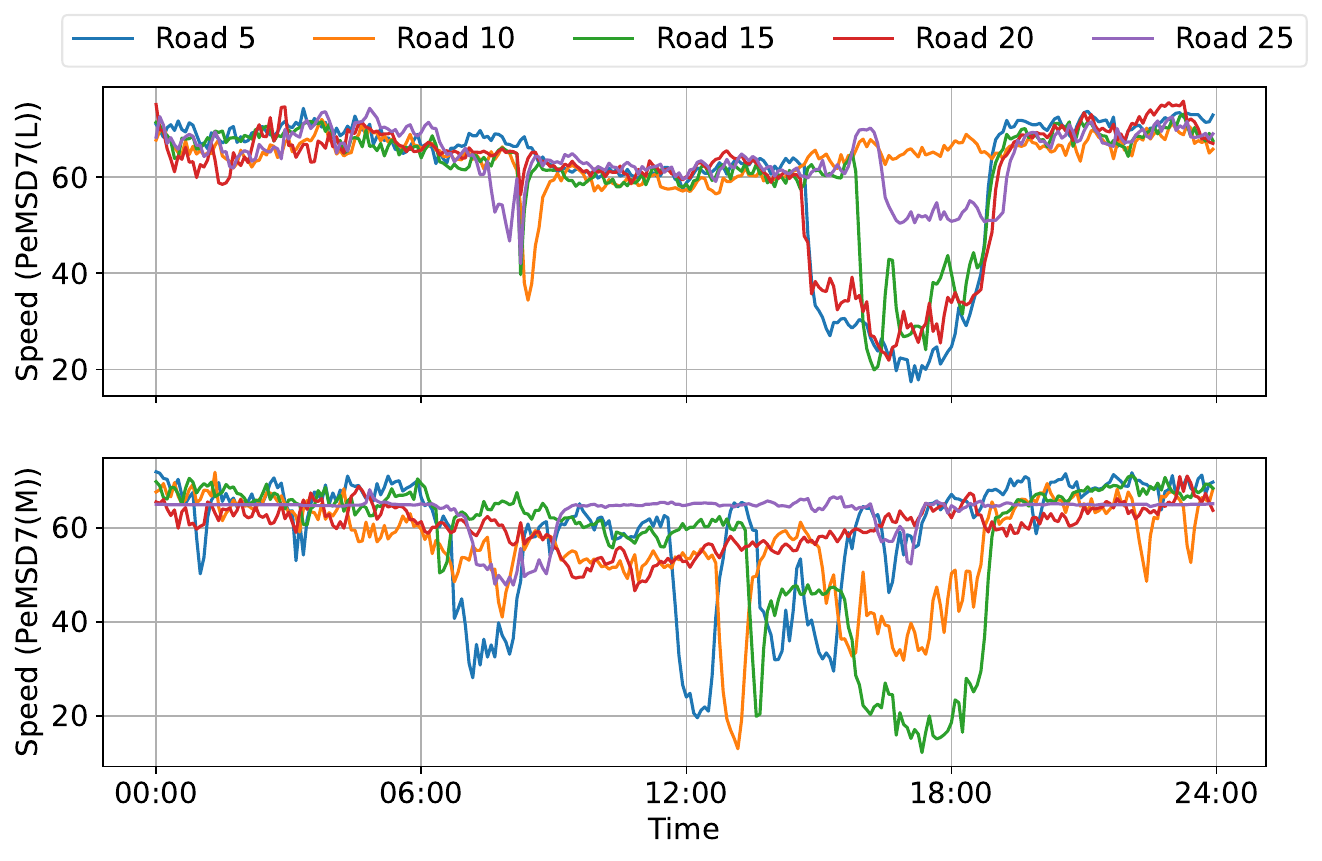}
        \caption{Speed data for five roads in the PeMSD7(L) and PeMSD7(M) datasets. The speed values fluctuate approximately between 20 and 80 km/h.}
        \label{fig: fig8}
\end{figure} 
 \begin{table}[h]
	\caption{Speed prediction results} 
	\centering
	\setlength{\tabcolsep}{3pt}
		\begin{tabular}{l ccccccc}
			\toprule
			\multicolumn{1}{l}{} & \multicolumn{3}{c}{PeMSD7(M)} & & \multicolumn{3}{c} {PeMSD7(L)}\\ 
			\cmidrule{2-4}\cmidrule{6-8}
			 Model & MAE & RMSE & MAPE & & MAE & RMSE & MAPE \\
			\midrule
                STGCN & 4.01 & 7.55 & 9.67 & & 4.84 & 8.28 & 11.76 \\
                STSGCN & 3.01 & 5.93 & 7.55 & & 3.61 & 6.88 & 9.13 \\
                GWNET & 3.19 & 6.24 & 8.02 & & 3.75 & 7.09 & 9.41 \\
                LSGCN & 3.05 & 5.98 & 7.62 & & 3.49 & 6.55 & 8.77 \\
                STFGNN & 2.90 & 5.79 & 7.23 & & 2.99 & 6.91 & 7.69 \\
                \addlinespace
                DCRNN & 3.83 & 7.18 & 9.81 & & 4.33 & 8.33 & 11.41 \\ 
                STGNN & 3.27 & 6.39 & 8.33 & & 3.72 & 7.11 & 9.21 \\
                TGCN & 5.98 & 9.97 & 16.91 & & 7.25 & 11.71 & 22.20 \\
                AGCRN & 2.79 & 5.54 & 7.02 & & 2.99 & 5.92 & 7.59 \\
                Z-GCNETs & \underline{2.75} & 5.62 & \underline{6.89} & & \underline{2.91} & \underline{5.83} & \underline{7.33} \\
              \addlinespace
              ASTGCN & 3.61 & 6.87 & 8.84 & & 4.09 & 7.64 & 10.25 \\
              MSTGCN & 3.54 & 6.14 & 9.00 & & 3.58 & 6.43 & 9.01 \\
              STG2Seq & 3.48 & 6.51 & 8.95 & & 3.78 & 7.12 & 9.50 \\
              DSANet & 3.52 & 6.98 & 8.78 & & 3.66 & 7.20 & 9.01 \\
              DSTAGNN & 2.78 & \underline{5.53} & 6.93 & & 3.03 & 6.04 & 7.65 \\
              \addlinespace
              STGODE & 2.97 & 5.66 & 7.36 & & 3.22 & 5.98 & 7.94 \\
              
              \midrule
             MAGCRN (Ours) & \textbf{2.68} & \textbf{5.43} & \textbf{6.71} & & \textbf{2.85} & \textbf{5.78} & \textbf{7.14} \\
                \addlinespace
                \bottomrule
	\end{tabular}
        \label{tab:table4}
\end{table}
\subsection{Results on Speed Prediction}

TABLE \ref{tab:table4} reports the average speed prediction results for the next 12 steps on PeMSD7(M) and PeMSD7(L) datasets. Similar to the flow datasets, MAGCRN outperforms all the baselines on the speed data. However, two different observations can be made from this experiment: 1) the performance gain on speed data is slightly lower than that on the flow data, and 2) an almost identical performance gain is obtained on both speed datasets even though the number of nodes is largely different. 

The former observation can be justified by the fact that the speed data generally differs from flow data in terms of node patterns, i.e., speed data has consistent periodic patterns and a lower fluctuation range than the flow data. This can be visually observed by comparing Fig. \ref{fig: fig8} of speed data with Fig. \ref{fig: fig1} of flow data.  The latter observation can be justified by the fact that both PeMSD7(M) and PeMSD7(L) fluctuate approximately in a similar range. Hence, the number of nodes becomes less influential as their growing number does not introduce major fluctuations across the time steps.  We believe the nature of speed data makes the prediction task easier than the flow data. Thus, the MAGCRN operations become of average worth to the speed prediction while highly valuable to the flow prediction task.
\section{Conclusions}
In this paper, we propose an effective GCRN-based predictor MAGCRN for traffic forecasting. In MAGCRN, we employ a GCRN as a backbone module and propose two novel modules, NMPL and NAWG to operate over the GCRN module. The former module utilizes a meta-learning technique hypernetwork to generate convolutional filter weights from the node parameters learned during GCRN operation. These convolutional filters are then convolved with the output of the last GCRN layer to extract node-specific features. This process allows MAGCRN to handle the recurrence limitation of local patterns by enriching node parameters with node-specific patterns extracted globally for each individual node. The latter module leverages a multihead cross-attention mechanism over the full sequence output of the GCRN module to generate attention weights for the node-specific features learned in the former module. These weights allow enriching node-specific features with short- and long-range patterns to handle the recurrence limitation of losing the previously learned features.  

Experiments on six real-world traffic datasets demonstrate that MAGCRN has superior predictive performance compared to a large pool of recent state-of-the-art traffic predictors. Besides, these experiments reveal several interesting observations, among which the most significant are summarized as follows:
\begin{itemize}
    \item MAGCRN reduces errors on most nodes in a given traffic flow network.
    \item MAGCRN improves the predictive performance more on flow datasets with a large number of nodes than the ones with a small number.
    \item MAGCRN improves predictive performance on both traffic flow and speed data.
    \item MAGCRN improves predictive performance on both the short- and long-term prediction tasks.  
    \item Both NMPL and NAWG have significant individual contributions to the overall predictive performance of the MAGCRN.
    \item MAGCRN is not highly sensitive to the network configuration and has a moderate computational cost.
\end{itemize}
In the future, we intend to implement an auto-regressive version of the MAGCRN and extend it to other time-series forecasting domains, such as economics, healthcare, finance, and climate. 
\bibliographystyle{IEEEtran}
\bibliography{IEEEabrv,ref}
\end{document}